\documentclass{article}
\usepackage{natbib}
\usepackage{graphicx}
\usepackage{amsfonts}
\usepackage{bibentry}
\usepackage{float}
\usepackage{amssymb}
\usepackage{amsmath}
\usepackage{subfigure}
\usepackage[margin=1.25in]{geometry}
\usepackage{makecell}
\usepackage{mathtools}
\usepackage{placeins}
\usepackage{color}
\usepackage{soul}
\usepackage{booktabs}
\usepackage{color,soul}

\newcommand{\argmin}[1]{\underset{#1}{\operatorname{arg}\,\operatorname{min}}\;}
\newcommand{\norm}[1]{\left\lVert#1\right\rVert}

\usepackage[inline]{enumitem}
\title{DeepRacing: Parameterized Trajectories\\ for Autonomous Racing}
\author{
  Trent Weiss\\
  Department of Computer Science \\
  University of Virginia\\
  \texttt{ttw2xk@virginia.edu} 
  \and
  Madhur Behl\\
  Department of Computer Science \\
  University of Virginia\\
  \texttt{madhur.behl@virginia.edu}
}
\begin{document}
\maketitle

\begin{abstract}
We consider the challenging problem of high speed autonomous racing in a realistic Formula One environment. 
DeepRacing is a novel end-to-end framework, and a virtual testbed for training and evaluating algorithms for autonomous racing. 
The virtual testbed is implemented using the realistic F1 series of video games, developed by Codemasters\textcopyright, which many Formula One drivers use for training. This virtual testbed is released under an open-source license both as a standalone C++ API and as a binding to the popular Robot Operating System 2 (ROS2) framework.  This open-source API allows anyone to use the high fidelity physics and photo-realistic capabilities of the F1 game as a simulator, and without hacking any game engine code. 
We use this framework to evaluate several neural network methodologies for autonomous racing. Specifically, we consider several fully end-to-end models that directly predict steering and acceleration commands for an autonomous race car as well as a model that predicts a list of waypoints to follow in the car's local coordinate system, with the task of selecting a steering/throttle angle left to a classical control algorithm. We also present a novel method of autonomous racing by training a deep neural network to predict a parameterized representation of a trajectory rather than a list of waypoints. We evaluate these models performance in our open-source simulator and show that trajectory prediction far outperforms end-to-end driving. Additionally, we show that open-loop performance for an end-to-end model, i.e. root-mean-square error for a model's predicted control values, does \emph{not} necessarily correlate with increased driving performance in the closed-loop sense, i.e. actual ability to race around a track. Finally, we show that our proposed model of parameterized trajectory prediction outperforms both end-to-end control and waypoint prediction.

\end{abstract}

\section{Introduction}
\label{sec:intro}
Vision-based solutions are believed to be a promising direction for autonomous driving due to their low sensor cost, and recent developments in deep learning. 
End-to-end models for autonomous driving have attracted much research interest ~\citep{DBLP:journals/corr/SantanaH16,DBLP:journals/corr/JanaiGBG17,DBLP:journals/corr/XuGYD16}, because they eliminate the tedious process of feature engineering. Algorithms for end-to-end driving are being trained and evaluated in both simulation~\citep{perot2017end,DeepDriving}.  These methods show promise and in some cases are implemented on real vehicles~\citep{bojarski2016end}. However, there is room for significant progress as these studies primarily use simulators with simplified graphics and physics ~\citep{wymann2000torcs,brown2018udacity} and so the obtained driving results lack realism. 
Additionally, there is little work showing how these methods will perform under ``edge" or ``corner" cases.  That is, highly unusual or unexpected scenarios where an autonomous agent must make a decision in an environment it has either not seen or wasn't specifically programmed to handle.  These cases are the cause of many failures in autonomous driving agents \citep{tesla,uber}.  There is an unfilled need to directly address these edge cases with autonomy that can be \emph{agile} and is capable of handling these unexpected edge cases safely.  As a first step toward agile autonomy, we consider the project of high-speed autonomous racing, specifically Formula One, as a proxy for agile behavior.   This approach is not unprecedented.  In the early 20\textsuperscript{th} century, a time of great consternation about giving up horses in favor of motorcars, Ford Motor Company used motorsport racing as a means of demonstrating to the public that it's products were safe for everyday users.  Additionally, many of the now-standard features of commercially available automobiles were conceived as innovations in motorsport racing.

Demonstrating high-speed autonomous racing can be considered as a grand challenge for vision based end-to-end models.
Autonomous racing can be considered an extreme version of the self-driving car problem, making progress here has the potential to enable breakthroughs in agile and safe autonomy.
To succeed at racing, an autonomous vehicle is required to perform both precise steering and throttle maneuvers in a physically-complex, uncertain environment, and by executing a series of high-frequency decisions.
Autonomous racing is also highly likely to become a futuristic motorsport featuring a head-to-head complex battle of algorithms~\cite{scacchi2018autonomous}. 
For instance, Roborace~\citep{roborace} is the Formula E's sister series, which will feature fully autonomous race cars in the near future. 
Autonomous racing competitions, such as F1/10 racing and Autonomous Formula SAE~\citep{koppula2017learning} are, both figuratively and literally, getting a lot of traction and becoming proving grounds for testing perception, planing, and control algorithms at high speeds. 

We present DeepRacing, a novel end-to-end framework for training and evaluating algorithms specifically for autonomous racing. 
DeepRacing uses the Formula One (F1) Codemasters game as a virtual testbed. This game is highly realistic - both in physics and graphics - and is used by many real world F1 drivers for training. 
Our DeepRacing C++ API enables easy generation of training data under a variety of racing environments, without the cost and risk of a physical racecar, and racetrack. 
This allows anyone to use the high fidelity physics and photo-realistic capabilities of the F1 game as a simulator, and without hacking any game engine code. 
The DeepRacing framework is open-source \textit{https://github.com/linklab-uva/deepracing}.

In addition we present AdmiralNet - a novel neural network architecture for predicting a parameterized trajectory (specifically, B\'ezier Curves) for the autonomous agent to follow. We conduct a case study of this model against several other autonomous racing approaches.  Our evaluation demonstrates the ability to train and test end-to-end autonomous racing algorithms using this highly photo-realistic video game environment.

\subsection{Contributions of this paper}
The paper has the following contributions:
\begin{enumerate}[noitemsep,nolistsep]
\item This is the first paper to demonstrate and enable the use of a highly photo-realistic Formula 1 Codemasters\textsuperscript{\textcopyright} game, with a high fidelity physics engine, as a test-bed for developing autonomous racing algorithms, and testing them in a closed-loop manner. 
\item We implement and evaluate a deep neural network (DNN) called AdmiralNet. AdmiralNet uses a novel combination of Convolutional Neural Networks (CNNs), Recurrent Neural Networks (RNNs), and B\'ezier Curves to autonomously race.
\item We compare AdmiralNet with several other approaches: both fully end-to-end (direct steering/acceleration prediction) and waypoint prediction in our closed-loop F1 testbed
\end{enumerate}

Section \ref{sec:problem_statement} provides a description of the problem and it's mathematical formulation.  Section \ref{sec:related_work} describes some related work in this domain.  Section \ref{sec:testbed} describes our open-source testbed and evaluation framework.  Section \ref{sec:method} describes our proposed network architecture and it's underlying mathematical basis in B\'ezier Curves.  Section \ref{sec:experiments} presents our empirical evaluation of both the open-source testbed and a case study of several autonomous racing methods within that testbed.  Section \ref{sec:conclusion} concludes and proposes some future work.



\section{Problem Statement:}
\label{sec:problem_statement}

The problem of autonomous driving distills to the task of scene understanding through sensor measurements, e.g. cameras, LIDAR point clouds, ultrasonic sensors, etc., and producing control inputs for the car, typically steering angle and throttle.  
Expressed mathematically, if the domain of the vehicle's entire sensor suite is $\mathbb{X}$ and the space of the vehicle's control outputs is $\mathbb{U}$, then the general problem of autonomous driving is a mapping from $\mathbb{X} \rightarrow \mathbb{U}$. 
Autonomous racing requires a great many control inputs: steering, acceleration, clutch, fuel mix selector, clutch bite point, and regenerative braking; just to name a few.  
For simplicity in this preliminary work, we assume that the control domain of the car is steering and acceleration.  
It is also assumed that these control values are linearly independent. 
This means the car's control domain is a 2-dimensional euclidean space, $\mathbb{R}^2$.  
An autonomous racecar would likely have multiple sensing modalities such as 2D vision, 3D LIDAR, and sound measurement.  
However, for this work, we focus on a vision based approach and assume that the car's input sensor domain is fixed-width images: $\mathbb{R}^{3xHxW}$, i.e. images with $3$ channels (3-channel color images are assumed for this work) of height $H$ and width $W$. 


We consider three high-level approaches to this problem:

\begin{enumerate}
    \item[1.] Fully End-To-End: training a network to map images directly to control inputs for the autonomous car (steering, throttle, etc.)
    \item[2.] Trajectory Waypoint Prediction: training a network to map sensor data to a series of waypoints for a car to follow, with a traditional controller such as Pure Pursuit deciding on control values for the car to best follow those waypoints
    \item[3.] Trajectory Prediction with B\`ezier Curves: similar to the second approach, but training a network to map sensor data to a parameterized representation of the trajectory itself rather than a sample of points on that trajectory.
\end{enumerate}

\begin{figure}
\centering
\includegraphics[width=1.0\columnwidth]{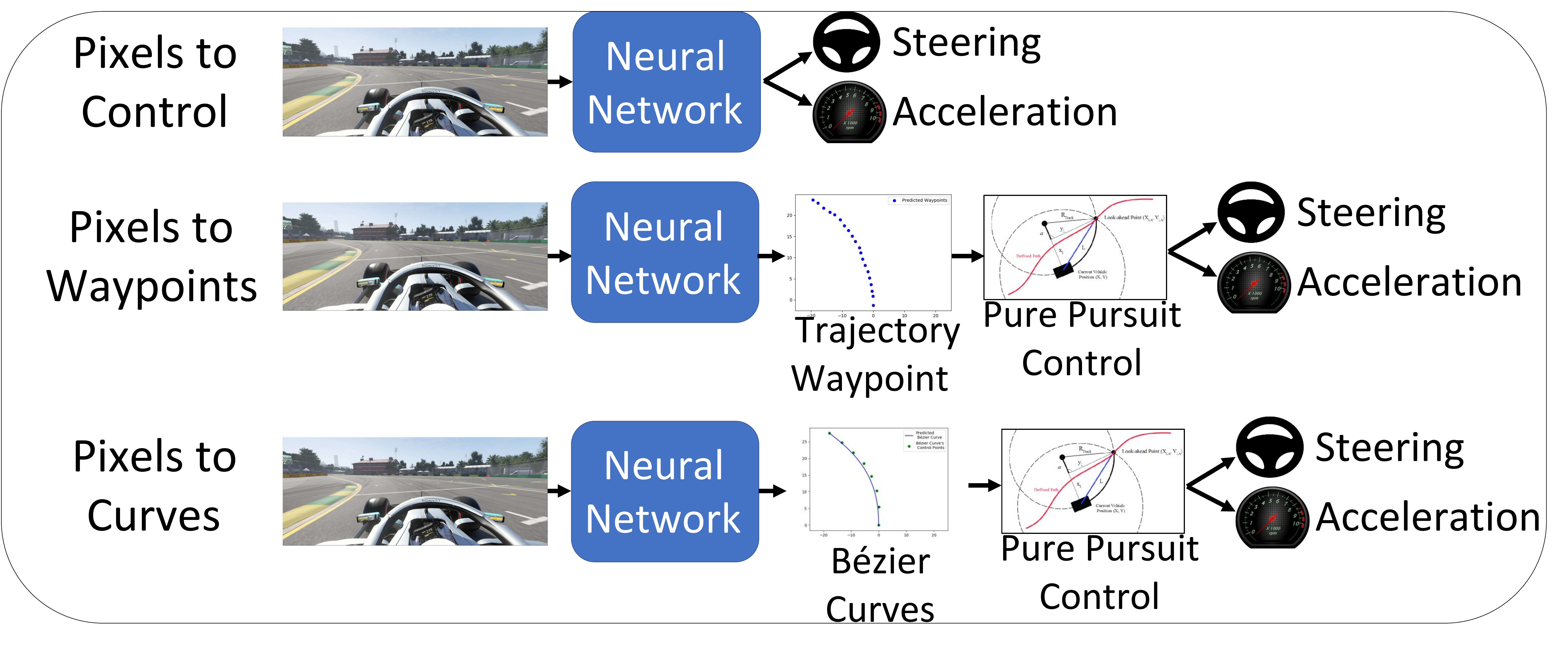}
\caption{Several approaches to autonomous racing.  Ultimately, any control strategy must provide steering and throttle values for the vehicle.  We present a case study of these approaches in section \ref{sec:experiments}.}
\label{fig:problem_approaches}
\end{figure}

Figure \ref{fig:problem_approaches} provides a graphical description of these three approaches.

\subsection{End-To-End Driving (Pixels to Control)}

There is a great body of work in this domain centered around the special case where $\mathbb{U}$ consists of only steering angles.  NVIDIA's PilotNet architecture~\citep{pilotnet} is considered a seminal work on this approach.  . It focuses on mapping what the car's sensor suite is seeing at the present time to a single control command, at the present time. This is done in an end-to-end manner i.e. the DNN is trained to directly map pixels to control outputs: $\mathbb{R}^{3 x H x W} \rightarrow \mathbb{R}^2$.

\begin{figure*}[!htb]
\centering
\hfill
\subfigure[Two similar images]{\includegraphics[width=0.3\textwidth]{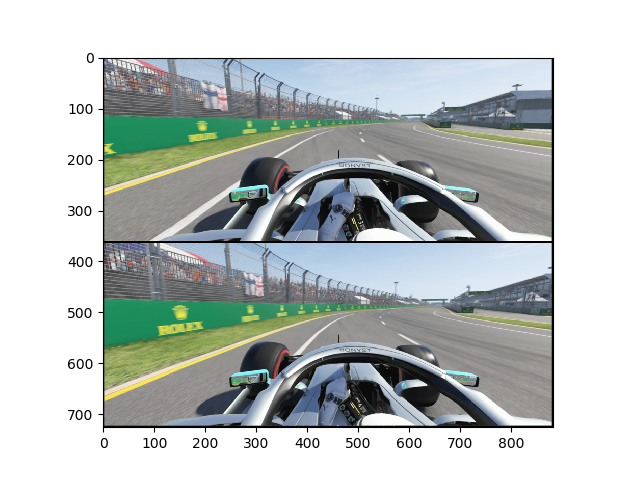}}
\subfigure[Corresponding trajectories]{\includegraphics[width=0.3\textwidth]{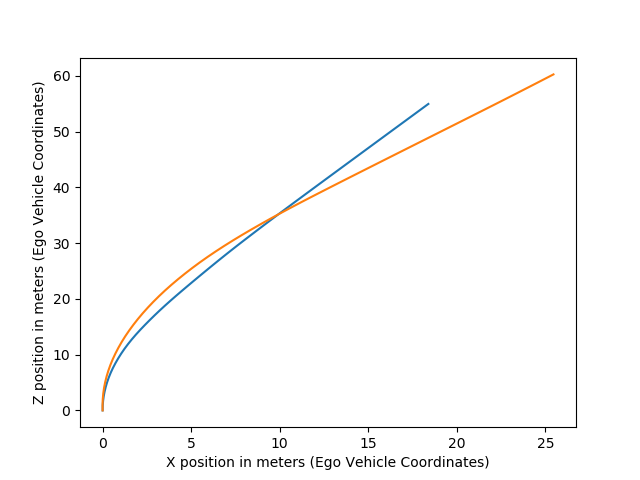}}
\subfigure[Distance between the two trajectories]{\includegraphics[width=0.3\textwidth]{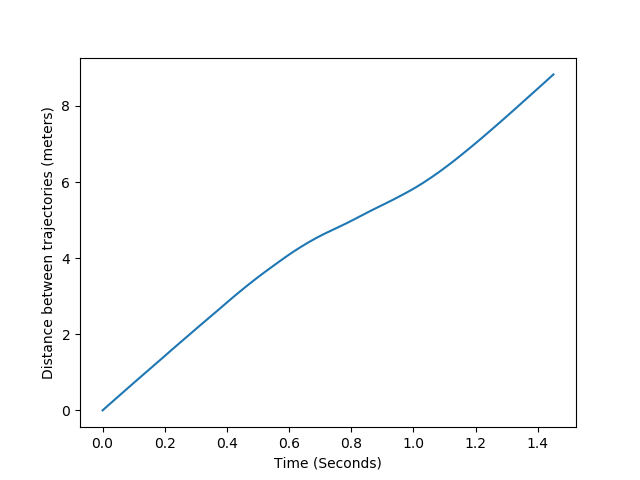}}
\caption{Purely Markovian methods like CNNs can present an ill-posed problem. For two very similar static images, the trajectories differ significantly.}
\label{fig:illposed}
\end{figure*}

However, this model of autonomous driving does a poor job of capturing how expert drivers behave.
For instance, a Formula One racing driver does not simply analyze the pixels directly in front of him and map those pixels directly to a single steering angle and throttle pressure.  An expert driver considers a history of previous observations to build up some temporal \emph{context} about the scene.  Additionally, using only single static images can create an ill-posed problem in which the map from pixels to control is not a function.  Consider the images and corresponding ground-truth trajectories in Figure \ref{fig:illposed}.  Even though the images are similar, their ground-truth trajectories differ noticeably. Within only 1.4 seconds, the paths diverge by almost 9 meters.  In a high-speed scenario like racing, this could easily be the difference between a successful race and a devastating crash.

For example, if a driver an ego vehicle sees a car directly in front of him but knows (from previous observation) that this car is moving away from him, he would behave differently than he would if he saw the exact same car in the exact same position, but with a relative velocity \emph{towards} the ego vehicle.  This problem of the exact same scene mapping to multiple control decisions, depending on context, does not lend itself to machine learning techniques that assume a functional mapping between input measurements and output predictions.  An alternative approach~\citep{deepracing-date} is to consider a sensor reading as a sequence of images: $\mathbf{I}_{i-N}, \mathbf{I}_{i-N+1}, \mathbf{I}_{i-N+2}, ..., \mathbf{I}_{i}$, where the subscript $i$ represents the current time and $c$ represents some number of time-steps into the past, as a single measurement that is then used to predict a sequence of control outputs ${\mathbf{r}}_{i+1}, {\mathbf{r}}_{i+2},{\mathbf{r}}_{i+3}, ... {\mathbf{r}}_{i+P} \in \mathbb{R}^3$, where ${\mathbf{r}}_{k}$  represents a control output at time $k$.  This view of the problem is a mapping from a \emph{context window} of sensor readings to an \emph{intent window} of control outputs for the autonomous vehicle:
\begin{equation}
\mathbb{R}^{N x C x H x W} \rightarrow \mathbb{R}^{P x 3}
\end{equation}

However, this fully end-to-end approach has it's weaknesses.  For very high-speed systems such as a race car, small errors in a control input map to very large errors in the actual path the car follows and can create catastrophic results.  This problem is somewhat intuitive to human drivers.  For example, when driving at highway speeds, human drivers prefer not to make aggressive steering maneuvers unless needed to avoid a crash. As opposed to low-speed situations, where quick changes in steering are quite common to make sharp turns, maneuver around an obstacle, or park a vehicle.

\subsection{Trajectory Prediction - Pixels to Waypoints}

To remedy this problem, one could view the problem of autonomous driving not as a static function from pixels to control, but as a temporally varying task that maps 1-dimensional manifolds, \emph{curves}, in a space of sensor inputs to curves in the ambient task space of the ego vehicle.   In this approach, rather than mapping a sequence of context images directly to a sequence of control outputs, the autonomous car needs to predict a sequence of \emph{waypoints} in the ego vehicle's task space.  I.e. the problem becomes mapping as sequence of images: $\mathbf{I}_{i-N}, \mathbf{I}_{i-N+1}, \mathbf{I}_{i-N+2}, ..., \mathbf{I}_{i}$ to a sequence of waypoints: $\vec{r}_{i+1}, \vec{r}_{i+2},\vec{r}_{i+3}, ... \vec{r}_{i+P}  \in \mathbb{R}^D$ where $D$ is the dimensionality of the task space.  Expressed mathematically, a function of the form:
\begin{equation}
\mathbb{R}^{N x 3 x H x W} \rightarrow \mathbb{R}^{P\text{ }x\text{ }D}
\end{equation}
For this work, we consider the sensor space as only images and the ambient task space is assumed to be a 2-dimensional manifold embedded in $\mathbb{R}^3$. This approach falls along a similar lines as end-to-end control by mapping a context window to a long-term intent, but the intent window is a set of points for the car to follow rather than a control schedule.  The task of mapping a predicted trajectory for the car to follow down to a specific steering and throttle value for the car's low-level control is left to classical control techniques based on a bicycle model of the car's kinematics.   For this work, a Pure Pursuit controller is used, but this model is generalizable to other control strategies.  

This approach also has it's weaknesses.  It suffers from the so-called ``curse of dimensionality" and can produce very non-smooth paths.  With so many parameters required to represent a single output ($P\text{ }x\text{ }D$ of them), models that are already prone to over-fitting like deep neural networks can predict very noisy output trajectories that can vary significantly with only minor (possibly imperceptible) changes in the input images.  Additionally, because these parameters represent \emph{waypoints}, they are not just a set of real numbers, their \emph{ordering} has specific and actionable meaning.  If a model predicts even a single waypoint incorrectly, this can cause a serious error in the car's chosen control action.  This is especially true for our chosen control algorithm of Pure Pursuit, if the incorrectly predicted point happens to be chosen as the lookahead point. Section \ref{sec:pure_pursuit} describes what this means in more detail.

\subsection{Trajectory Prediction - Pixels to Bezier Curves}

Finally, we consider a novel approach to autonomous racing. Rather than training a neural network fully end-to-end, we view the problem of trajectory prediction not as a mapping from images to waypoints, but from images to a parameterized description of a smooth 1-manifold embedded in the car's task space.  B-Splines are a very intuitive choice, as they are $C^\infty$ curves and are commonly used for motion planning tasks~\citep{bsplineplanning,bsplineplanning2}. However, their recursive representation does not fit well with gradient back-propagation, making it difficult to apply them in the context of machine learning.  
To remedy this, we consider using \emph{B\'ezier Curves} as a canonical form of curves in an autonomous racecar's task space.  
B\'ezier Curves are a linear combination of Bernstein Polynomials and are described in more detail in section \ref{sec:bezier_curves}.

Regardless of the chosen approach. Any supervised machine learning method technique requires significant amounts of training data for fitting a model.  
We describe our fully open-source DeepRacing testbed and data collection infrastructure for gather training data in a highly photorealistic racing environment in Section~[\ref{sec:testbed}].

\section{Related Work}
\label{sec:related_work}

There is existing literature for end-to-end autonomous driving. We divide the related work into simulation testbeds and autonomous driving methods and provide a brief reprise on both.

\subsection{Autonomous driving simulators}

There exists a great deal of related work in simulation for autonomous driving (AD). Traditionally, simulation capabilities have been primarily used in the planning and control phase of AD~\citep{likhachev2009planning,buehler2009darpa,katrakazas2015real,anderson2010optimal,best2017autonovi}.
More recently, simulation has been used in the entire AD pipeline, from perception and planning to control~\cite{pendleton2017perception}.
Waymo has claimed that its autonomous vehicle has been tested for billions of miles in their proprietary simulation system, CarCraft~\cite{madrigal2017inside}, little technical detail has been released to the public in terms of its fidelity for training machine learning methods. Researchers have tried to use images from video games to train deep-learning-based perception systems~\cite{johnson2016driving,richter2016playing}.

There are several examples of video games being used as a simulator to aid development of autonomous driving.   
Several are based on the popular Grand Theft Auto (GTA) game~\citep{playingfordata}, utilizing the high fidelity graphics of Rockstar Games'\textsuperscript{\textcopyright} state-of-the-art rendering engine.  
However, while being photo-realistic, this game is known for flaws in the underlying physics engine, since it was not intended to be used as a simulator. This technique also requires a modification to the underlying game engine code in order to extract data (steering, acceleration, etc.) attached to each game screenshot.  This modification ran afoul of Rockstar's copyright protections, and both of these projects infamously received cease and desist letters from Rockstar Games to pull the code from public domain.  
While creating DeepRacing, we ensured that our F1 2019 simulator requires no such ``hacking" of the unaderlying game engine. It uses public APIs for screen capturing as well as a stream of UDP packets that the F1 game broadcasts.
End-to-end driving was showcased in the car racing game TORCS~\citep{wymann2000torcs} using Reinforcement Learning but its physics and graphics lack realism.
The data collected from the racing game TORCS, for example, have a biased distribution in visual background and road traffic and thus severely diverge from real driving scenarios.
Microsoft AirSim~\citep{airsim} and CARLA~\citep{dosovitskiy2017carla} are examples of open-source autonomous driving simulators but they are are largely restricted to urban driving scenarios and are not suited for development and testing of end-to-end autonomous racing.

\subsection{Autonomous Driving Architectures: Fully End To End}
In one of the earliest work on end-to-end autonomous driving NVIDIA~\citep{pilotnet} presented the PilotNet CNN architecture. PilotNet is a feed-forward style network that directly regresses to a single steering value for each input image obtained from a front facing dashboard camera.
However, PilotNet is limited by it's inability to capture temporal information.  
Each input image is run through the CNN separately with no time-varying context around that image. This is not just a limitation of PilotNet, but of CNNs in general.  
~\cite{event-frames} remedy this problem  with an event camera to batch images from an arbitrary number of time-steps as an input to a CNN and achieve a performance increase.
\cite{memory-cells} present a different approach that uses Long Short-term Memory Cells~\citep{LSTM} as a means of capturing a history of the steering trajectory and encode temporal structure of the problem.  
\cite{discrete-action-model} present a similar technique that uses a Fully Convolutional Network (FCN) to extract a feature representation of the input space, but limit their model to classification among a discrete set of actions: go straight, stop, left turn, and right turn. 

~\cite{auto-encoder} use a novel combination of CNN and a traditional auto-encoder approach. This network uses a CNN for feature extraction and applies an encoding function to translate the regression problem into a more manageable classification problem.  
Eraqi also presents the novel concept of a ``sliding window", a variable length temporal sequence that is input into the LSTM, allowing the model to encode temporal information at arbitrary lengths. 
While their results indicate that auto-encoding the LSTM outputs improves performance over directly regressing to steering angle, their model also only predicts steering one time-step into the future and does not capture the expert driver's long-term intent.  

~\cite{DeepDriving} also present a novel approach that blends expert domain knowledge of highway driving by defining a notion of image affordance that is then mapped to a steering command.  However, like PilotNet, their approach only considers current image data and is limited to a fixed set of affordance templates that are purpose-built for only highway driving.

\subsection{Autonomous Driving Architectures: Pixels to Trajectories}

However, end-to-end control has it's limits.  Wrapping the entirety of a perception-planning-control pipeline into one black-box network can exacerbate the the problem of overfitting, as a single deep network needs to learn all three high-level tasks as part of a single model.  We show in our experiments in section \ref{sec:experiments} that just because an end-to-end (pixels $\longrightarrow$ control) method performs well on unseen validation data (as measured Root-Mean-Squared error), that does not mean that same method contains a useful driving strategy that will perform well in an actual live-driving test.  

To help remedy this problem of over-fitting and produce more scalable methods, there is existing work in autonomous driving by trajectory prediction (pixels $\longrightarrow$ trajectories). For example, \cite{LSTM-highway} use an LSTM cell to predict a series of waypoints for images of highway driving at varying traffic densities. ~\cite{deep-path-planning} present an approach for generating waypoints with a CNN and then using a Model Predictive Controller (MPC) to control the vehicle. ~\cite{learning-by-cheating} use a similar approach to predict waypoints based on expert demonstration.  \cite{ChauffeurNet} is a blend of several techniques, but their neural network is penalized (increased loss) for veering too far off of a prescribed list of waypoints.

\subsection{Autonomous Driving Architectures: Other Approaches}
\cite{LSTM-occupancy-grid} use a similar approach but predict an occupancy grid of the ego vehicle's immediate surroundings. This occupancy grid is then used as input for a classical motion planning approach to determine control actions. \cite{pan2017virtual} use a Reinforcement Learning (RL) approach to autonomously drive in a TORCS simulator.  However, their experimental evaluation is limited to an open-loop accuracy evaluation on unseen test data and does not offer closed loop testing to evaluate real driving performance. \cite{high-speed-RL} use a similar approach and also intentionally focus on high-speed scenarios, similar to racing.

\section{DeepRacing: F1 Racing Simulation}
\label{sec:testbed}
In order to train an end-to-end neural network to race autonomously using the context and the intent described in the previous section; we need a reliable way to generate annotated training data. 
Obtaining such annotated data for real motor-sport racing drivers is difficult since these data are often trade-secrets and not available in the public domain. 
\begin{figure*}
\centering
\includegraphics[width=0.9\linewidth]{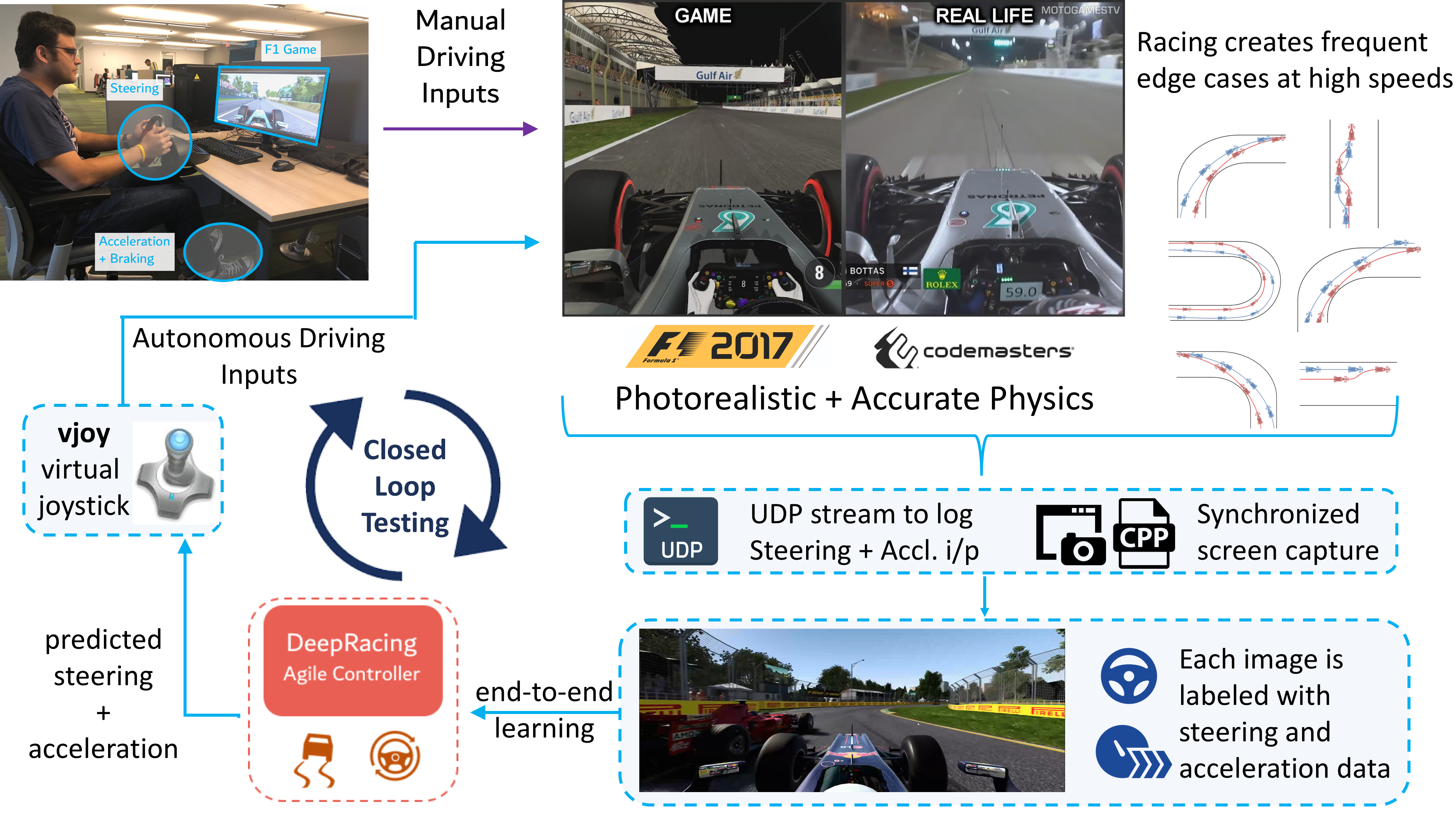}
\caption{We use the F1 Codemasters\textsuperscript{\textcopyright} game as a virtual testbed for training and closed-loop testing for our autonomous racing deep neural network. This is the first time the highly photo-realistic and high fidelity physics engine enabled game has been used as a CPS testbed in an entirely legal way without cracking and tweaking game engine code.}
\label{fig:deepracing}
\end{figure*}

Furthermore, it is not enough to only obtain training data, but also important to close the loop and autonomously race in the same environment to enable evaluation of end-to-end models, and reinforcement learning approaches. Consequentially, we use the F1 2019 racing game released by Codemasters\textsuperscript{\textcopyright} as a realistic testbed for autonomous racing that is built on top of decades of work in 3D graphics and physics simulation.  We now describe the novel features this first-of-it's kind testbed provides.

\subsection{DeepRacing realism}
\noindent \textbf{Photo-realism and physics modeling}: The game is extremely photo-realistic, as shown in Figure~\ref{fig:deepracing}, and is based on high-fidelity simulated Newtonian physics, with detailed simulation of the car's drive train, high-speed aerodynamic drag, and even a simulated traction/slip interface between the car and the track surface. Due to its realism, the F1 series was the first game to be used in the Formula One eSports Series, which debuted in 2017~\citep{esport}. There is also evidence to suggest that real-life F1 drivers use this game for practicing~\citep{max}. The photo-realism in the driver's point-of-view combined with the physics realism of the game's engine provide a strong opportunity to gather training data as close to real-world racing scenarios as one can get without the cost and risk of a real race-car.  

\noindent \textbf{Real weather having real effects}: The racecar reacts to weather conditions, in terms of tire degrading, braking, and handling. For example, under wet track conditions braking at the last minute before a corner, the car will have a different response as opposed to braking under dry conditions.
Rainstorms are intense, reducing visibility and slowing races down, and again, that’s replicated in this game.

\noindent \textbf{Deep customisation}: The game facilitates a high degree of customization including adjustable dead zones, linearity, and saturation for vehicle control.  Settings for aerodynamics, traction, tyre choices, etc. are also highly customizable. 

\noindent \textbf{Closed-Loop Testing}: Our infrastructure supports the ability to push steering and throttle commands back into the game.  This closed-loop support will enable AI researchers to develop and test autonomous racing policies with this highly realistic game environment. 


\noindent Figure \ref{fig:deepracing} gives a graphical description of this testbed.  
\subsection{UDP data stream}
The game advertises a ``fire-and-forget" data stream of telemetry data containing a variety of information about the game's current state over a User Datagram Protocol (UDP) network socket.  
Each packet in the stream is a snapshot of the game's state tagged with a timestamp for when that state was generated. 
A full description of the F1 telemetry stream and all of the information it provides is available on a Codemaster's\textsuperscript{\texttrademark} forum~\cite{f1_udp}. 
The state variables broadcast by the game include, but are not limited to:
\begin{enumerate}
    \item Steering angle, throttle and brake of all vehicles (including the ego vehicle)
    \item Position and velocity of all vehicles (including the ego vehicle)
    \item Various state information about the ego vehicle such as wheel speed, amount of fuel remaining, and tire pressure.
\end{enumerate} 
\subsection{DeepRacing framework}
Unfortunately, the game does not tag each packet of state information with an image of the driver's point-of-view at that packet's timestamp.  Because supervised machine-learning techniques require \emph{labeled} training data, we present the DeepRacing API, a fully open-source C++17 API for both grabbing screenshots of the driver's perspective (the ``ego" vehicle) in the F1 game and automatically tagging them with ground-truth values of the game's state at the time that image was captured.  
In this C++ framework, a single dedicated process spawns two threads:
\begin{enumerate}
\item For capturing screenshots of the drivers point of view, which we call the ``screen-capture thread''
\item For listening for telemetry data from the game on a UDP socket, which we call the ``telemetry thread''
\end{enumerate}
Each thread has a copy of a shared CPU timer. The screen-capture thread uses our C++17 API, built on top of Microsoft's DirectX, for capturing images of the driver's point of view and tags each image with a system timestamp from the operating system.  The telemetry thread listens for UDP data from the game and tags each packet with a system timestamp \emph{from the same system clock}.  

\subsubsection{Data Synchronization}
Because each UDP packet is also tagged with a session timestamp from the F1 game's internal (and black box) game clock, we can then convert OS timestamps to game session timestamps by fitting a least-squares regression line to the session timestamps versus the OS timestamps of the UDP packets.  Note that, if our infrastructure is functioning properly, this relationship must be a line with slope close to $1.0$, as 1 second on the game clock should correspond to exactly 1 second on the OS clock.  We use the $r^2$ value of this regression line as well as how close its slope is to $1.0$ to evaluate how closely the timescale in our logging infrastructure matches the internal clock of the F1 game.   Results from this evaluation are included in section \ref{sec:experiments}.  

The UDP broadcast operates at 60 Hz and our screencapture framework can achieve frame-rates of up to 35 Hz, enabling high-fidelity data-logging.

\subsection{Open-loop testing}
\label{subsec:openloop}
\begin{figure}[!htb]
\centering
\includegraphics[width=1.0\textwidth]{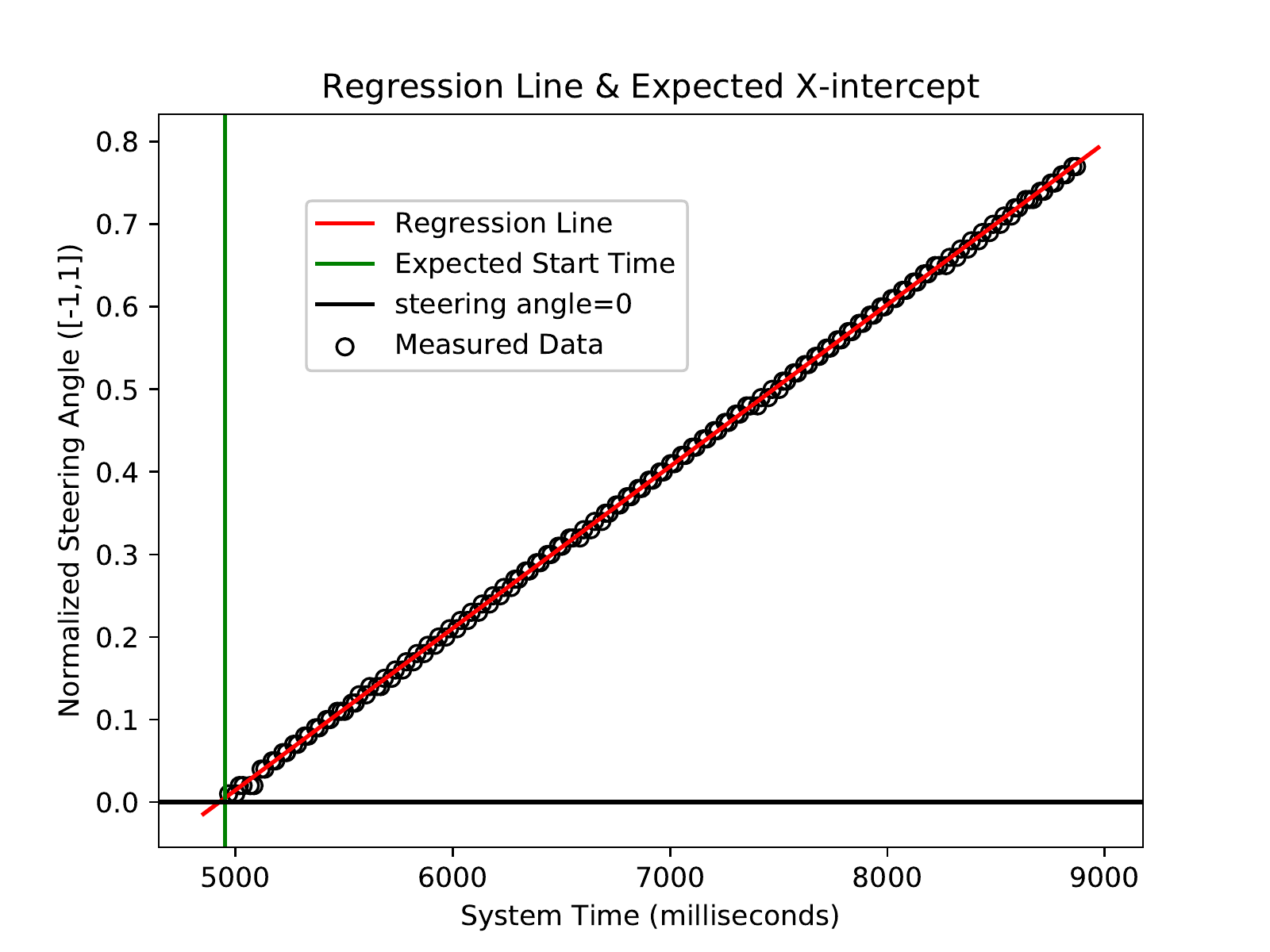}
\caption{An example experiment evaluating our virtual joystick infrastructure.  For this experiment, the ramp-up in steering angle began at a system time of 4.952950 seconds.  The regression line shown has an x-intercept of 4.926211 seconds, indicating a latency of $26.79\text{ milliseconds}$ }
\label{fig:lagtest}
\end{figure}

Once each image is given a game session timestamp, state information from the UDP stream can then be assigned to each image by interpolation over the relevant state variable with respect to session time.  For example, in our experiments, we use B-Spline fitting to determine the ego vehicles position and spherical linear interpolation over unit quaternions to determine the ego vehicle's orientation as a function of session time.  We also evaluate our ability to push steering and acceleration control inputs back into the game via a virtual joystick API built on top of vJoy~\citep{vjoy} by measuring the latency between when a command was applied to the virtual joystick and when that command is reflected in the game's UDP stream.  This experiment was conducted by setting the steering angle on our virtual joystick to 0.0 at a known system timestamp, then linearly ramping it up to 1.0 (all the way to the left) at a fixed update rate.  We then measure the latency of our system by fitting a regression line to the measured steering angles coming off the UDP stream versus their respective system timestamps.  If this system has 0 latency, the x-intercept (time) of this regression line would be the exact system time the ramp-up began.  If there is some latency, then the x-intercept will be offset from the ramp start.  An example of such a regression experiment is plotted in figure \ref{fig:lagtest}.  Our experiment shows an x-intercept of 4926.16 milliseconds for a ramp-up that was begun at a system time of 4952.95 milliseconds.  This indicates a lagtime of $4952.95 - 4926.16 = 26.79\text{ milliseconds}$ between when control commands are applied and when they are reflected in the game.  However, the UDP stream only publishes at $60Hz$, which means as much as $\frac{1}{60Hz} = 16.\bar{6}\text{ milliseconds}$ of that delay could stem from the UDP stream itself.  In other other words, it's possible that as much as $16.\bar{6}\text{ milliseconds}$ of the measured lagtime stems from the fact that changes in the game's current state might not get published on the UDP stream for as much as $16.\bar{6}\text{ milliseconds}$.

\subsection{Closed-loop testing}
\begin{figure}[!htb]

\centering
\subfigure[Racelines]{\includegraphics[width=0.48\textwidth]{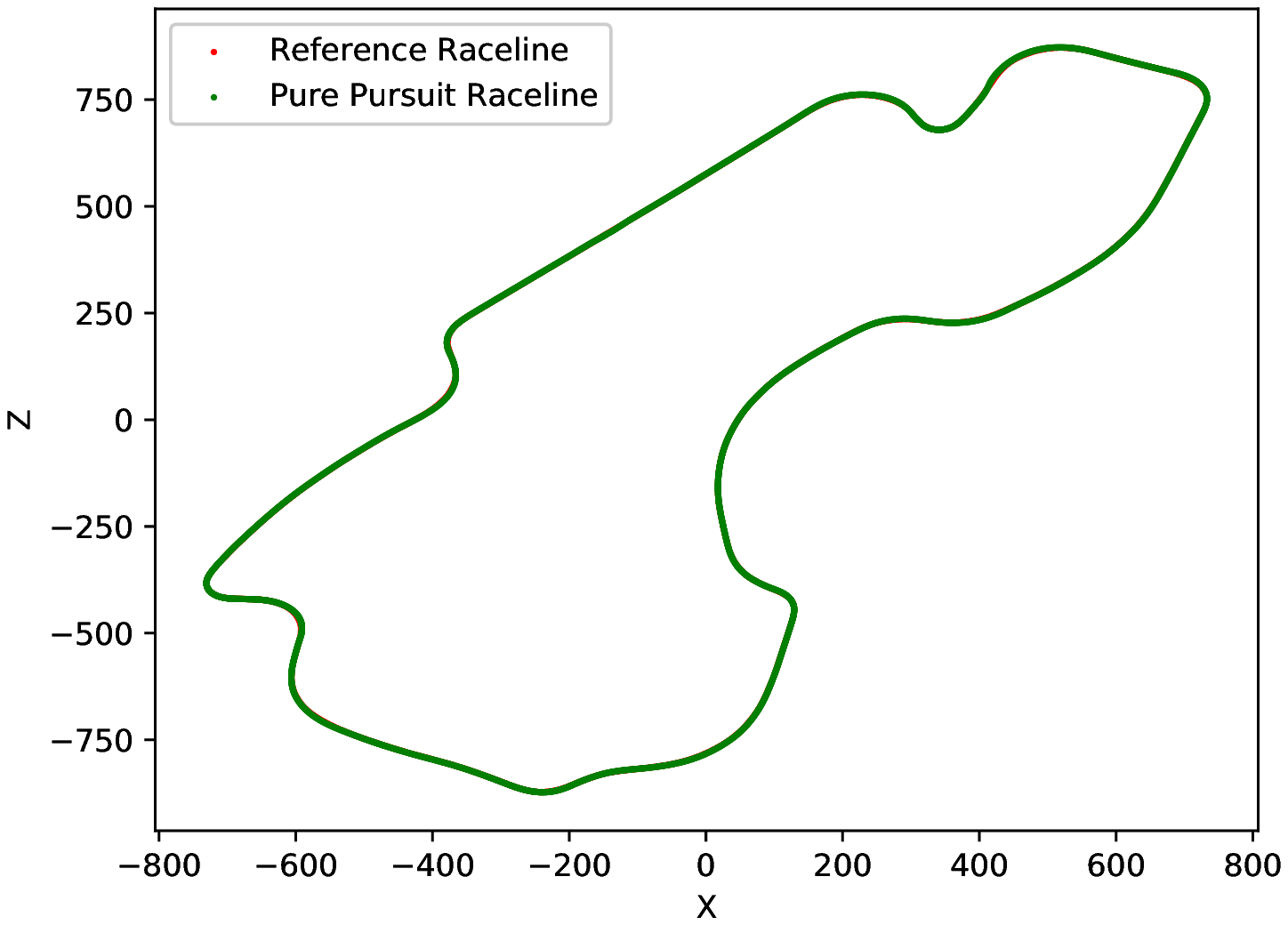}}
\subfigure[Distance Histogram]{\includegraphics[width=0.48\textwidth]{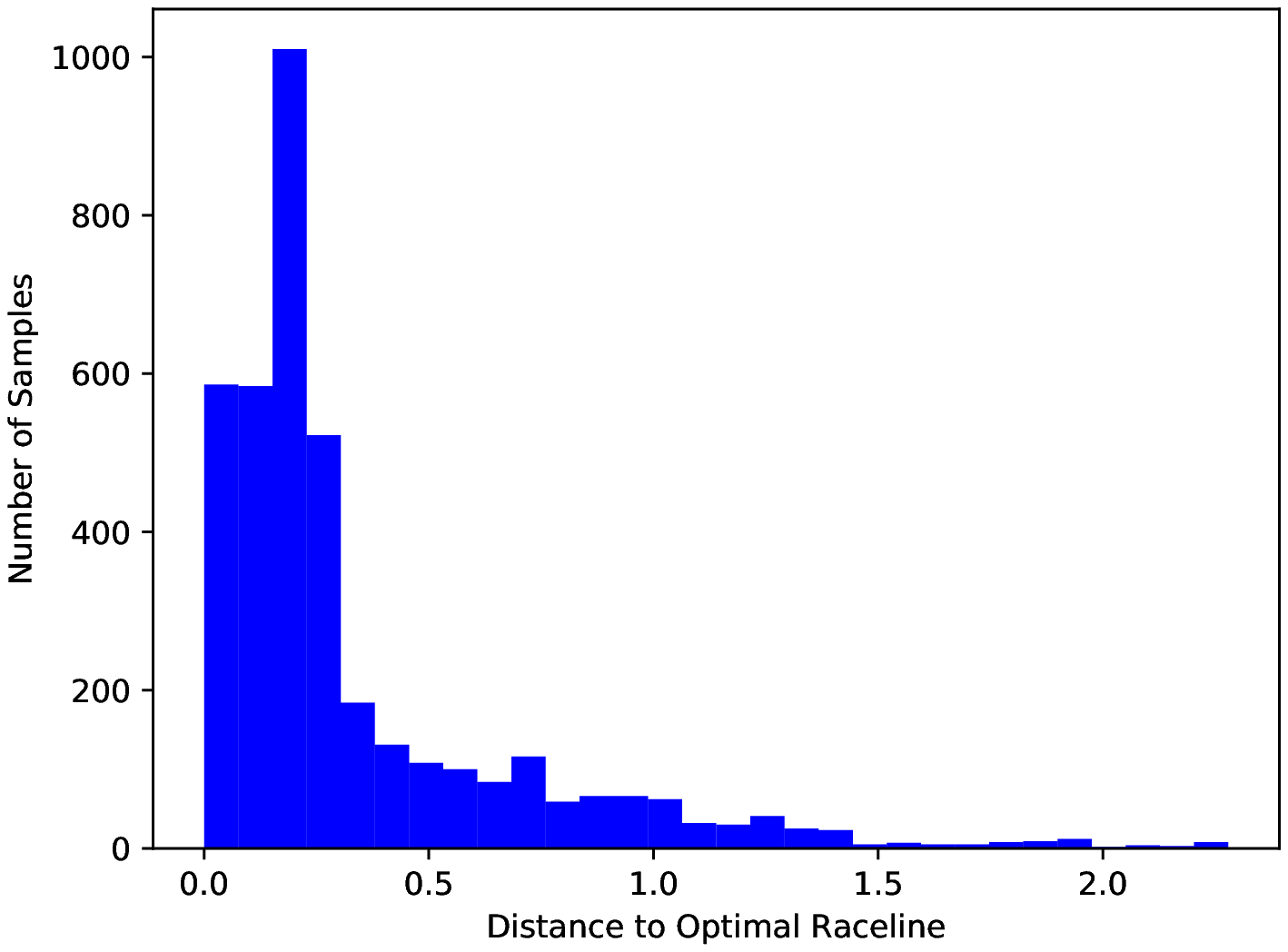}}

\caption{A comparison between the path followed by our closed-loop infrastructure an the optimal raceline for the Australia Circuit. The two paths are almost indistinguishable.}
\label{fig:pure_pursuit_test}
\end{figure}

Finally, our test-bed setup also supports the ability to close the loop and autonomously drive the F1 car in the game using control inputs predicted by autonomous driving policies.  This is accomplished with the same virtual joystick setup described in subsection \ref{subsec:openloop}.
We evaluate this closed-loop capability by utilizing ``oracle" data in the form of a pre-recorded list of vehicle positions (waypoints) that represent the optimal raceline for the Australia circuit in the F1 game and a simple pure-pursuit controller to steer the car with ground-truth knowledge of the track's optimal raceline. We then measure the effectiveness of this controller by the distance from the path followed by the pure-pursuit controller to this optimal raceline.  Note that this evaluation is \textbf{not} intended to evaluate any particular autonomous driving model, but is intended to show the accuracy of our closed-loop framework for testing such models.  
This closed-loop test shows a mean distance to the optimal raceline of \textbf{0.340936 meters}. Figure \ref{fig:pure_pursuit_test} a) shows an optimal raceline overlaid with the path followed by our pure pursuit control, the two paths are almost indistinguishable. Figure \ref{fig:pure_pursuit_test} b) is a histogram of the distance between the pure pursuit control's path and the optimal raceline. 

\subsection{DeepRacing API}
This data collection and testing infrastructure, implemented in C++, is called the DeepRacing API, the first closed-loop environment of it's kind for collecting training data, and testing learned models on simulated F1 race-cars in the photo-realistic F1 2019 game.  The software itself is architected as an object-oriented library that exposes a simple interface for allowing user-written code to handle data captured by the underlying infrastructure.

A simple interface, called IF1FrameGrabHandler in the C++ API,  is exposed that defines what the library does with new timestamped images as they comes off of the screencapture, with the ``hard work" of actually obtaining that data hidden in a separate library.  User's need only overwrite the appropriate methods in this interface with an implementation of what to do with captured images.  A similar interface, called IF1DatagrabHandler in the C++ API, is exposed for handling captured UDP packets.  Again, users need only overwrite a few methods to receive timestamped UDP packets as a C++ object. This clean abstraction allows users to define whatever data flow model makes sense for their application. To generate datasets for our experiments in \ref{sec:experiments}, we implement this interface to save image data and UDP data to the file system with a buffered approach to ensure no packets or images are lost during the delay inherent to filesystem access.  Other data flow models are possible, e.g. sending data to a cloud-based server, displaying to screen, etc.   Creating such models within our framework is only a matter of implementing the simple C++ interfaces. 
 
\subsection{DeepRacing ROS2 Interface:}
 Additionally, we present bindings to the very popular Robot Operating System 2.0 (ROS2). A ROS2 Publisher class is provided that broadcasts both telemetry data and images from the F1 game on several ROS2topics (fire-and-forget broadcasts).  Any ROS2 nodes can then listen to these data streams, enabling researchers to integrate this testbed with existing codebase in a ROS2 ecosystem.  All of this software has been released under an Open-Source license.
\begin{figure}[!htb]
\centering
\includegraphics[width=0.9\columnwidth]{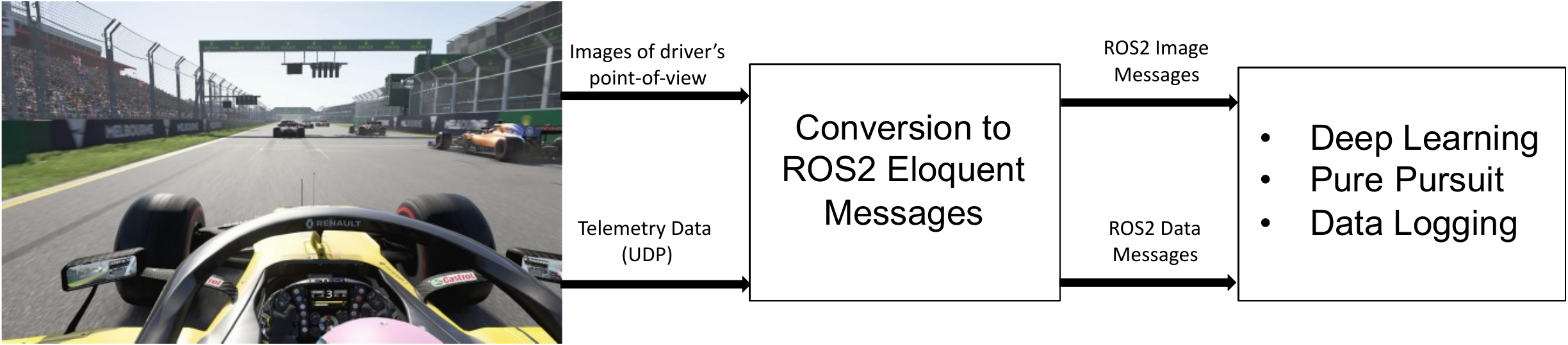}
\caption{ROS2 integration for the DeepRacing framework. In this figure, ``Image Transport" refers to an open-source ROS2 library that handles compressing/decompressing images and routing them through the ROS2 network via a set of rostopics. F1 2019\texttrademark is a registered trademark of Codemasters\textcopyright. The Formula One logo is a registered trademark of the Formula One Group.}
\label{fig:ros2_integration}
\end{figure}

%
%

%
%

\section{AdmiralNet: Mapping Context to Intent}
\label{sec:method}



The second research contribution of this work is  AdmiralNet, an improvement over NVIDIA's PilotNet~\cite{pilotnet}, a deep neural network capable of learning to race autonomously. 
We train two forms of AdmiralNet, one designed to map images to waypoints for a car to follow, and another to map images to a parameterized trajectory for a car to follow, in our case: B\'ezier Curves.
The key building blocks in AdmiralNet's architecture are:
\begin{enumerate}
    \item A Pure Pursuit low-level controller.
    \item 2D Convolution Neural Networks
    \item 3D Convolution Neural Networks
    \item Recurrent Neural Networks
    \item B\'ezier Curves
\end{enumerate}

We go over each of these components of our implementation next.


\subsection{Pure Pursuit Control}
\label{sec:pure_pursuit}
\begin{figure}[!htb]
\centering
\includegraphics[width=.5\columnwidth]{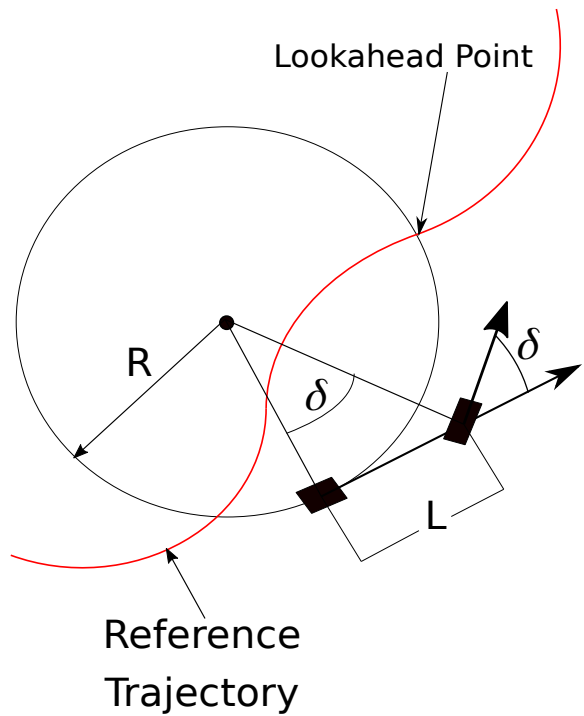}
\caption{Illustration of Pure Pursuit Control. A steering angle $\delta$ is selected such that a car with wheelbase (distance between the axles) L follows a circular arc towards a lookahead point selected from a reference trajectory. The radius of this circle, R, can vary as it needs to and becomes infinite in the case of the car directly facing the lookahead point}
\label{fig:pure_pursuit}
\end{figure}

Pure pursuit, first described in \cite{pure_pursuit}, is a widely used algorithm for path following in front-wheel steered, non-holonomic vehicles~\citep{underwater_pure_pursuit,review_pure_pursuit}.  At it's core, given a sequence of waypoints $\{W_1, W_2, W_3, ... W_n \in \mathbb{R}^d\}$ expressed in the car's local coordinate system, Pure Pursuit sets a steering angle that puts the car on the circular arc connecting it's current position and a desired ``lookahead-point"(see Figure \ref{fig:pure_pursuit}).  
A common practice is to select the point that is closest to a \emph{lookahead distance} away from the ego vehicle, with the lookahead distance set to a constant factor times the ego vehicle's current speed.  That is:

$$d_{lookahead} = \gamma v$$
$$W_{lookahead} = \argmin{W}{ \vert \norm{W} -d_{lookahead} \vert}$$

Where $v$ is the ego vehicle's current speed and $\gamma$ is a tuneable parameter.  This proportional control strategy keeps the car from making maneuvers that are too aggressive for the car's actuators to handle at the expense of potentially ``cutting corners" on sudden sharp turns at high speed.  

As will be described later, in our novel AdmiralNet framework, we produce a set of waypoints for the autonomous racecar to follow, and the low-level Pure-Pursuit controller will generate the steering and velocity commands to follow those waypoints. 

We evaluate the closed-loop performance of our pure-pursuit implementation in Section~\ref{sec:experiments}.

\newpage

\subsection{B\'ezier Curves}
\label{sec:bezier_curves}
A B\'ezier curve is a parametric curve heavily used in computer graphics and related fields. 
The curve, a linear combination of Bernstein Polynomials, is named after Pierre B\'ezier, who (somewhat poetically) developed them to model car bodies on Renault racecars.

A B\'ezier curve is formed from a combination of Bernstein polynomials that maps a scalar parameter $t \in [0,1]$ to a point in a euclidean space of dimension $d$, $\mathbb{R}^d$.  More specifically, a B\'ezier Curve is a weighted combination of a set of ``control points", with the weights computed from the Bernstein polynomial basis.  These control points define a B\'ezier curve as follows~\citep{mathematics-for-cg}:

For a set of control points:
$$\mathbb{P}  = \{ \mathbf{P}_0,  \mathbf{P}_1, \mathbf{P}_2, \mathbf{P}_3, .. \mathbf{P}_n \in \mathbb{R}^d \} $$
The corresponding B\'ezier curve, $\mathbf{B} : [0,1] \rightarrow \mathbb{R}^d$, as a function of the parameter, $t$, is:
$$\mathbf{B}(t) = \sum_{k=0}^n\binom{n}{k}{(1-t)}^{n-k}t^k\mathbf{P}_k$$.
\begin{figure*}
\centering
\begin{subfigure}
  \centering
  \includegraphics[width=.6\columnwidth]{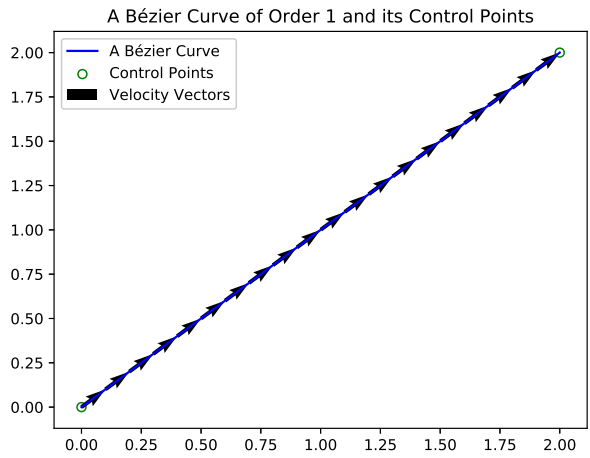}
\end{subfigure}
\begin{subfigure}
  \centering
  \includegraphics[width=.6\columnwidth]{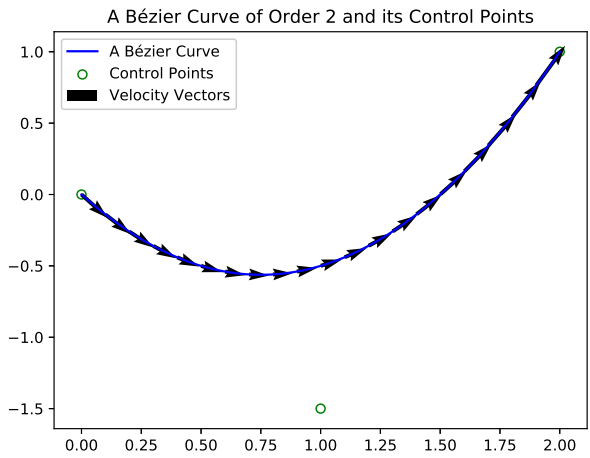}
\end{subfigure}
\begin{subfigure}
  \centering
  \includegraphics[width=.6\columnwidth]{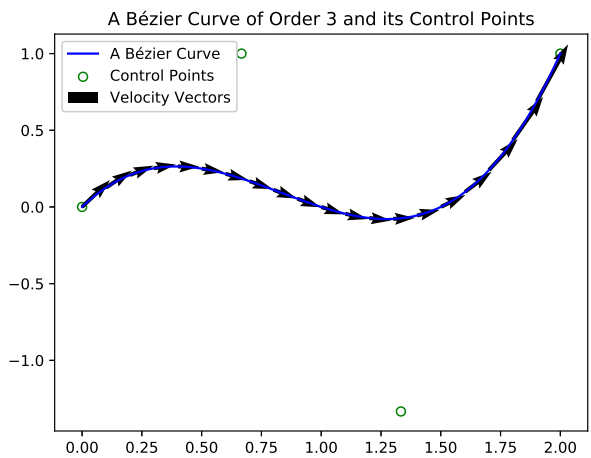}
\end{subfigure}
\caption{Several B\'ezier Curves of varying degree}
\label{fig:bezier_curve_figure}
\end{figure*}
Figure \ref{fig:bezier_curve_figure} shows several such curves.  For most practical cases, as well as our case, the parameter $t$ will represent time (or step size) normalized to the interval $[0,1]$

The dimensionality of the ambient euclidean space, $d$, is called the ``dimension" of the B\'ezier curve and the integer $n$, 1 less than the number of control points, is called the ``degree" or ``order" of the curve and is also the degree of the underlying Bernstein polynomials.  If the parameter, $t$, is normalized to the interval $[0,1]$, then the B\'ezier curve always starts at $\mathbf{P}_0$ and ends at $\mathbf{P}_n$, but does not necessarily pass through any of the control points from $\mathbf{P}_1$ to $\mathbf{P}_{n-1}$.

We present a method of evaluating B\'ezier curves and their derivatives framed as a \emph{matrix multiplcation}, making such a step very easy to integrate into a machine learning model trained by a gradient back-propagation algorithm.   Consider a fixed time-vector, $\mathbf{t}$ with $N$ elements linearly spaced on the interval $[0,1]$ i.e $\mathbf{t}_k = \frac{k}{N-1}$:

$$\mathbf{t} = {[0,\frac{1}{N-1},\frac{2}{N-1},\frac{3}{N-1}, ...,\frac{N-3}{N-1},\frac{N-2}{N-1} , 1 ]}$$

Consider a time matrix: $\mathbf{A}$ such that
$$\mathbf{A_{ij}}(\mathbf{t},n) = \binom{n}{j}{(1-\mathbf{t}_{i})}^{n-j}{\mathbf{t}_{i}}^{j}$$

$\mathbf{A}$ would then be $N$ rows by $n$ columns with each row corresponding to a time on the interval $[0,1]$ and each column corresponding to a term of the B\'ezier curve's Bernstein basis polynomial.

Consider a control point matrix, $\mathbf{P}$, such that the $k_{th}$ row of $\mathbf{P}$ is just the $k_{th}$ control point of the curve.
$$\mathbf{P} =
                    \begin{bmatrix}
                        \mathbf{P}_0 \\
                        \mathbf{P}_1 \\
                        \mathbf{P}_2 \\
                        ... \\
                        \mathbf{P}_{n-2} \\
                        \mathbf{P}_{n-1} \\
                        \mathbf{P}_{n}
                    \end{bmatrix}
$$
$\mathbf{P}$ would then be $n+1$ rows by $d$ columns, with each column corresponding to a dimension in the curves ambient euclidean space.

Evaluating a B\'ezier curve, $\mathbf{B}$, of order $n$ with a control point matrix $\mathbf{P}$ on all of the times in $\mathbf{t}$ then just becomes a matrix multiplication:

$$\mathbf{B}(\mathbf{t}) = \mathbf{A}(\mathbf{t},n)\mathbf{P}$$

Additionally,  B\'ezier curve's derivatives can also be evaluated as follows:
$$\mathbf{B^{'}}(t) = n\sum_{k=0}^{n-1}\binom{n-1}{k}{(1-t)}^{n-k-1}t^k(\mathbf{P}_{k+1}-\mathbf{P}_k)$$.

Note that this is equivalent to:

$$\mathbf{B^{'}}(t) = n\mathbf{A}(\mathbf{t},n-1)\Delta\mathbf{P}$$

Where $$\Delta\mathbf{P} =
                    \begin{bmatrix}
                        \mathbf{P}_1 - \mathbf{P}_0 \\
                        \mathbf{P}_2 - \mathbf{P}_1\\
                        \mathbf{P}_3 - \mathbf{P}_2 \\
                        ...\\
                        \mathbf{P}_{n}- \mathbf{P}_{n-1}
                    \end{bmatrix}
$$
And has 1 row fewer than $\mathbf{P}$.

We also present a method for fitting a bezier curve to a set of points (in the least-squares sense).  For the same time vector, $\mathbf{t}$, if given a matrix of points $\mathbf{L}$ sampled at the times in $\mathbf{t}$ such that:
$$\mathbf{L} =
                    \begin{bmatrix}
                        \mathbf{l}_0 \\
                        \mathbf{l}_1 \\
                        \mathbf{l}_2 \\
                        ... \\
                        \mathbf{l}_{N-2} \\
                        \mathbf{l}_{N-1} \\
                        \mathbf{l}_{N}
                    \end{bmatrix}
$$
 I.e., each $\mathbf{l}_k \in \mathbb{R}^d$ is a point sampled at time $\mathbf{t}_k$. The task of least squares fitting becomes:
 
 $$\mathbf{P}^{*} = \argmin{\mathbf{P}}\norm{\mathbf{A}(\mathbf{t})\mathbf{P} - \mathbf{L}}$$
 
 The solution to this least-squares problem can be framed in terms of the Singular Value Decomposition of $\mathbf{A}$.
 
 $$\mathbf{A} =\mathbf{U}\mathbf{\Sigma}\mathbf{V}^T$$
 
 $$\mathbf{P}^{*} = \mathbf{V}\mathbf{\Sigma}^{-1}\mathbf{U}^T\mathbf{L}  $$


We use B\'ezier curves as a parameterized representation of trajectories for a pure pursuit controller as well as a means of predicting trajectories for a racecar to follow in subsection \ref{sec:bezier_predictor}.


\subsection{Fully End-To-End Baseline}
\label{sec:fully_e2e}


A large body of work in this domain focuses on fully end-to-end algorithms\citep{3dconv-lstm,huston2008visuomotor,katzsupervised}.  
I.e. a network that is trained to map pixels in images directly to control outputs for the vehicle.  
In this paper, we consider two baseline algorithms in this domain:

\begin{enumerate}
    \item NVIDIA's end-to-end PilotNet CNN~\citep{pilotnet}
    \item A CNN-LSTM architecture that uses the output of a CNN as the input for a Long Short-Term Memory (LSTM) Cell.
\end{enumerate}

Both of these architectures are trained fully end-to-end under a loss function of the mean-squared-error between the ground truth steering and throttle for a particular image and the predicted steering and throttle:

$$
\mathbb{L}_{endtoend} = \frac{1}{2}[{(\phi^* - \hat{\phi})}^2 + {(\mathit{a}^* - \hat{\mathit{a}})}^2]
$$

Where $\phi^*$ and $\hat{\phi}$ are the ground-truth and predicted steering angles; and $\mathit{a}^*$ and $\hat{\mathit{a}}$ are the ground-truth and predicted throttle values for the car.  Although it is technically possible to apply both throttle and brake at the same time, doing so can severely damage a racecar's drive train. For this work, we only consider throttle as a single scalar value in which positive indicates acceleration and negative indicates braking.  We show in our experiments in section \ref{sec:experiments} that this approach to autonomous racing performs very poorly in a closed-loop sense.

\subsection{AdmiralNet For Waypoint Prediction}
\label{sec:waypoints}
We present a network architecture designed to map sequences of images (a context window) to sequence of wayppoints in the car's ambient task space (an intent window).   The input context window of this model is a sequence of $C$ color (RGB) images, each with height $H$ and width $W$, such that each input tensor is $Cx3xHxW$ (the second dimension is 3 because each RGB image has 3 channels).   For our experiments, we use $C=5$, $H=66$, and $W=200$.  It's output is a sequence of waypoints in the car's task space that are predicted to be optimal in some sense. For our experiments, we define ``optimal" to be the trajectory followed by an expert example upon seeing that sequence of images.  Additionally, we approximate the car's ambient task space as the euclidean plane, $\mathbb{R}^2$, of the car's axles. This architecture is named The AdmiralNet Waypoint Predictor.
\begin{figure}[!htb]
\includegraphics[width=\columnwidth]{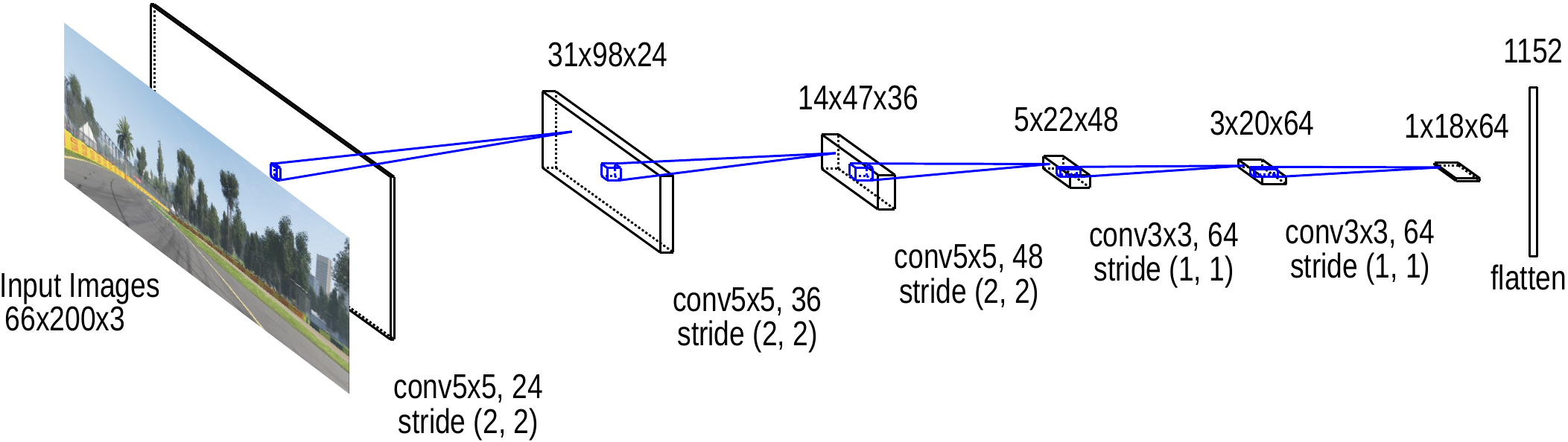}
\vspace{-5pt}
\caption{Our CNN that maps images to feature vectors. These feature vectors serve as a learned encoding of the context window. Although omitted from the graphic for clarity, batch normalization layers are also included after each convolutional layer. Image/feature map sizes are displayed as $\text{Height }x\text{ Width }x\text{ Channels}$
}
\label{fig:dcnn}
\end{figure}

\begin{figure}
\includegraphics[width=\columnwidth]{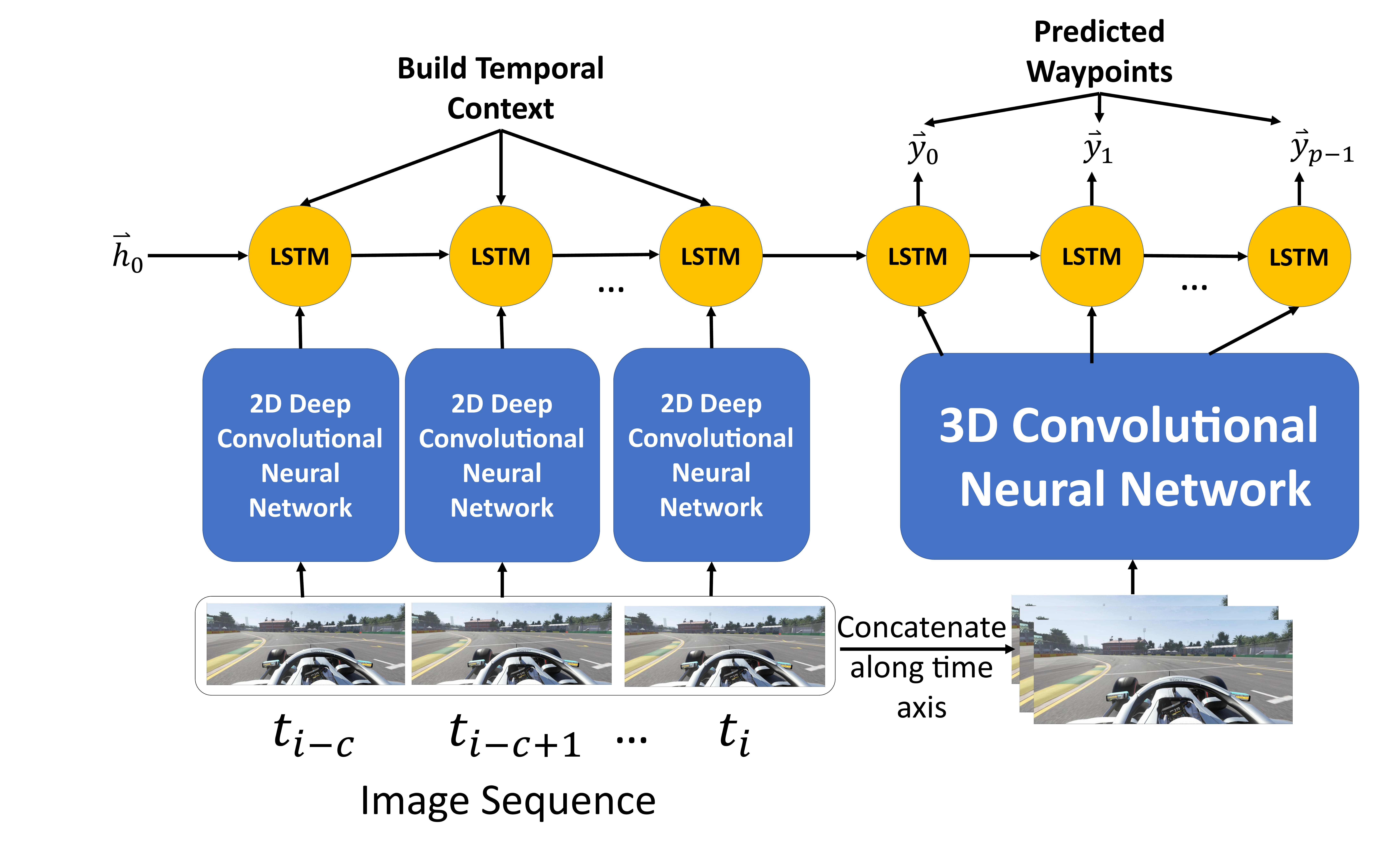}
\caption{Architecture For The AdmiralNet Waypoint Predictor. The Deep CNN layers are the ones shown in Figure~\ref{fig:dcnn} 
}
\label{fig:admnet_waypoint}
\end{figure}

Each image is passed through a Convolutional Neural Network (CNN) that maps each image to a ``deep feature" vector. These $C$ feature vectors are then used as the inputs to $C$ recurrent calls to an LSTM with a hidden dimension $h$ to build up the LSTM's hidden state with a learned encoding of the context window.  We also extend~\cite{3dconv-lstm}'s method of 3D ``spatio-temporal convolution" by passing the same sequence of images through a 3D convolutional network. The output of this 3D convolution is then used as the input for $p$ additional recurrent calls to the LSTM.  The resulting $p$ outputs are then passed to a linear layer with input dimension $h$ and output dimension 2. The outputs of this linear layer are taken as the sequence of predicted waypoints, $[ \hat{\Vec{y}}_1, \hat{\Vec{y}}_2, \hat{\Vec{y}}_3, ...\hat{\Vec{y}}_p  \in \mathbb{R}^2]$, each being a point in the car's local coordinate system, consisting of a lateral axis and a forward axis.  To produce control outputs of steering and throttle, these points are then passed to a Pure Pursuit Controller described in Section \ref{sec:pure_pursuit}.  For our experiments, we train this network to minimize the average euclidean distance between the predicted waypoints and the ground-truth waypoints, $[ \overset{*}{{\Vec{y}}_1}, \overset{*}{{\Vec{y}}_2}, \overset{*}{{\Vec{y}}_3}, ..., \overset{*}{{\Vec{y}}_p} \in \mathbb{R}^2]$, followed by a human example for that same sequence of images:

$$\mathbb{L}_{waypoint} = \sum_{i=0}^{p}\sqrt{\norm{\hat{\Vec{y}}_i - \overset{*}{{\Vec{y}}_i} } }$$
To enable an ablative analysis of this 3D convolutional subnetwork, we also test this same model with fixed constants used as the input for the final $p$ calls of the LSTM. These constants are learned parameters of the model and are optimized during the network training process.  The results of this ablative analysis are contained in section \ref{sec:experiments}.
\subsection{AdmiralNet For B\'{e}zier Curve Prediction}
\label{sec:bezier_predictor}

Waypoint prediction also has it's limits. Derivative information cannot be implicitly encoded in a list of waypoints, they must be inferred numerically and are therefore not well suited to gradient back-propogation.  Additionally, this approach suffers from the ``curse of dimensionality".   Each trajectory required $N x d$ parameters to represent, forcing the machine learned model to learn significantly more parameters to predict more dense trajectories.

To address these limitations, we present a novel approach for predicting future trajectories by using B\'{e}zier Curves as a dimensionality reduction technique.  This approach is very similar to the approach in Section \ref{sec:waypoints}, but we train our model to predict the control points of a B\'{e}zier Curve instead of directly predicting waypoints. I.e., \emph{we train a model to predict a parameterized representation of a curve rather than to learn samples from that curve}.  This method addresses both limitations of the waypoint method. Derivatives of a B\'{e}zier Curve can be readily computed in closed-form.  Additionally, the curse of dimensionality is mitigated by the fact that any number of points can be sampled from a B\'{e}zier Curve without the need to learn any additional parameters in a machine learned model.  We employ the same methodology as in Section \ref{sec:waypoints} to build up a temporal context for the LSTM and use some recurrent calls to project that context into an intent window.  However, we now employ two key additional techniques:

\begin{enumerate}
    \item The outputs of the $p$ additional calls to the LSTM are concatenated along the time axis to form a 2D ``feature grid". This feature grid is then interpreted as an image and passed through another CNN
    \item The output of this feature grid CNN is then linear mapped to predicted control points for a B\'{e}zier Curve rather than directly predicting waypoints
\end{enumerate}

\begin{figure}[!htb]
\includegraphics[width=\columnwidth]{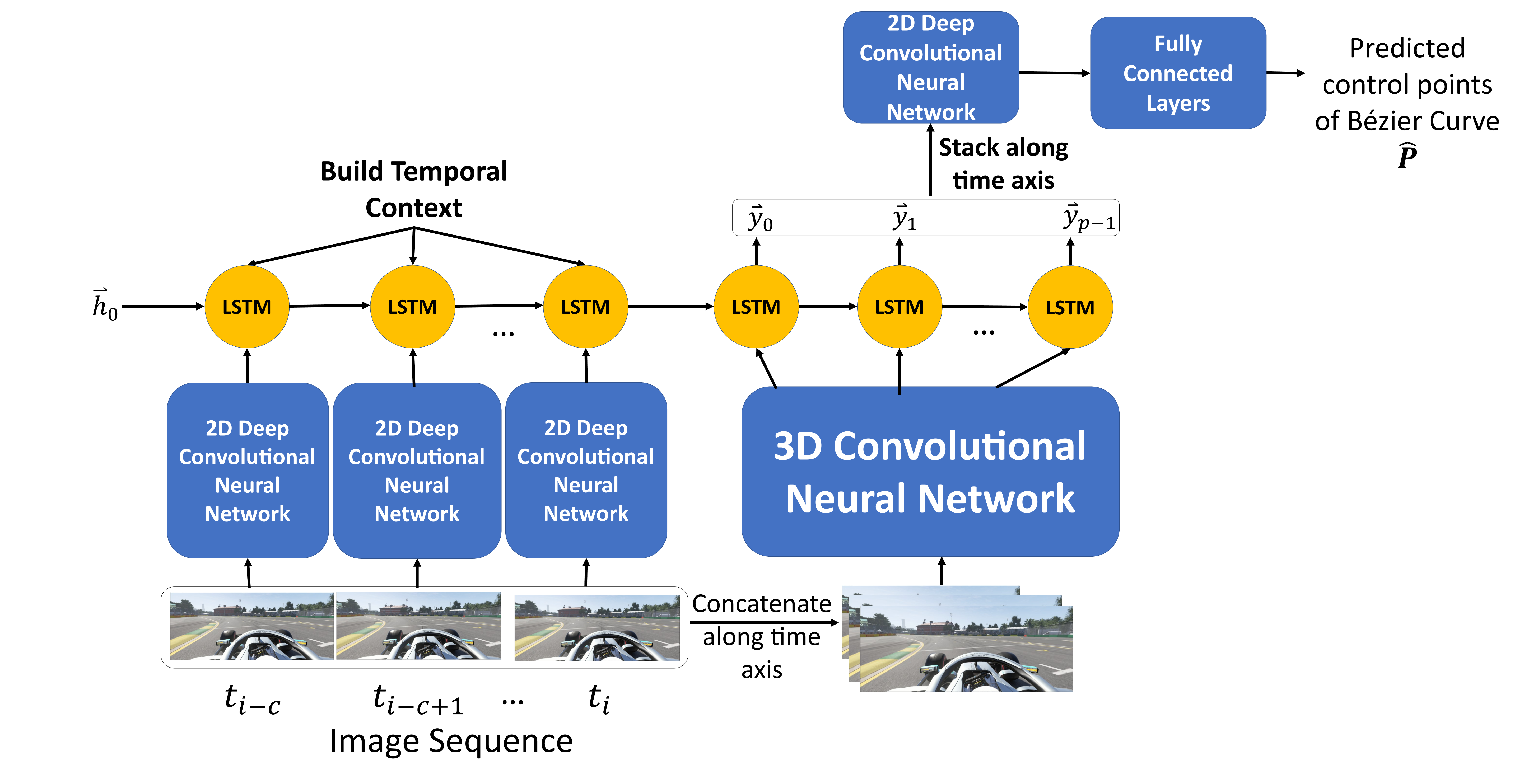}
\vspace{-5pt}
\caption{Architecture For The AdmiralNet B\'{e}zier Curve Predictor. The Deep CNN layers are the ones shown in Figure~\ref{fig:dcnn}}
\vspace{-20pt}
\label{fig:admnet_bezier}
\end{figure}

Just as for waypoint prediction, we use a Pure Pursuit controller to map a predicted trajectory to steering and throttle.  However, we evaluate the predicted B\'{e}zier Curve on a fixed sample of the interval $[0,1]$ to produce predicted waypoints rather than predicting the waypoints directly. For each pair of image sequences and ground truth waypoints, the network is trained to minimize a loss function that is a weighted sum of three terms:

\begin{enumerate}
    \item The average squared norm between the predicted B\'{e}zier Curve control points and that of a least squared fit (described in section \ref{sec:bezier_curves}) to the ground truth trajectory points: What we call ``control point loss"
    \item The average euclidean distance between the predicted B\'{e}zier Curve evaluated on the interval $[0,1]$ and the ground truth waypoints: What we call ``Position Loss"
    \item The average euclidean distance between the predicted B\'{e}zier Curve's derivative (velocity) evaluated on the interval $[0,1]$, but rescaled to the same time range as the ground truth labels, and the ground truth velocities: What we call ``Velocity Loss"
\end{enumerate}

One could also use sums instead of averages, but this only scales each loss term by a constant factor.  The quantities described above are computed as follows for B\'{e}zier Curves of degree $n$, i.e. with $n+1$ control points. $N$ represents the number of ground truth position/velocity waypoints sampled for each image sequence. $\hat{\mathbf{P}}$ is the matrix of control points predicted by AdmiralNet, $\overset{*}{{\mathbf{P}}}$ is the control point matrix derived from a least squares fit (from section \ref{sec:bezier_curves}) to the ground truth waypoints, $\hat{\mathbf{Y}}$ is the sequence of predicted waypoints obtained by evaluating $\hat{\mathbf{P}}$ on $[0,1]$.

\begin{equation}
    \mathbf{t} =[t_0,t_1,t_2,...t_{N-1}]
\end{equation}

\begin{equation}
\Delta t = t_{N-1} - t_0
\end{equation}

\begin{equation}
   \mathbf{s}= \frac{\mathbf{t} - t_0}{\Delta t} \subset [0,1] 
\end{equation}


\begin{equation}
   \hat{\mathbf{Y}} = [\hat{\Vec{y}}_0, \hat{\Vec{y}}_1, \hat{\Vec{y}}_2, ... , \hat{\Vec{y}}_{N-1} ]= \mathbf{A}(\mathbf{s}, n)\hat{\mathbf{P}}
\end{equation}


\begin{equation}
 \frac{\hat{d\mathbf{Y}}}{d\mathbf{s}} = n \mathbf{A}(\mathbf{s}, n-1)\Delta\hat{\mathbf{P}}
\end{equation}

\begin{equation}
 \frac{\hat{d\mathbf{Y}}}{d\mathbf{t}} = [\frac{d\hat{\Vec{y}}_0}{dt}, \frac{d\hat{\Vec{y}}_1}{dt}, \frac{d\hat{\Vec{y}}_2}{dt}, ... , \frac{d\hat{\Vec{y}}_{N-1}}{dt}] = \frac{\hat{d\mathbf{Y}}}{d\mathbf{s}} \frac{d\mathbf{s}}{d\mathbf{t}}
\end{equation}

\begin{equation}
 \frac{d\mathbf{s}}{d\mathbf{t}} =  \frac{1}{\Delta t} 
\end{equation}

\begin{equation}
 \frac{\hat{d\mathbf{Y}}}{d\mathbf{t}}= \frac{\hat{d\mathbf{Y}}}{d\mathbf{s}} \frac{1}{\Delta t}  
\end{equation}

\begin{equation}
\mathbb{L}_{position} = \frac{1}{N}\sum_{i=0}^{N-1}{\norm{\hat{\Vec{y}}_i - \overset{*}{{\Vec{y}}_i} } }
\end{equation}


\begin{equation}
\mathbb{L}_{velocity} = \frac{1}{N}\sum_{i=0}^{N-1}{\norm{\frac{d\hat{\Vec{y}}_i}{dt} - \overset{*}{{\frac{d\hat{\Vec{y}}_i}{dt}} } } }
\end{equation}

\begin{equation}
\mathbb{L}_{control\textunderscore point} = \frac{1}{n-1}\sum_{i=0}^{n}{\norm{\hat{\Vec{p}}_i - \overset{*}{{\Vec{p}}_i} } }
\end{equation}

The overall loss function for the AdmiralNet B\'{e}zier Curve Predictor is a weighted sum of these three loss functions:

\begin{equation}
\mathbb{L} = w_{position}\mathbb{L}_{position} + w_{velocity}\mathbb{L}_{velocity} + w_{control\textunderscore point}\mathbb{L}_{control\textunderscore point}
\end{equation}


Following the same procedure as with the Waypoint Predictor, we perform an abalative analysis of the additional 3D convolutional subnetwork by replacing it's output with learnable constants.  The results of this ablative analysis are contained in section \ref{sec:experiments}.
\subsection{Comments on Implementation Details}
\label{sec:implementation_comments}
Autonomous racing is a very high-speed task.  As such, a very fast running time for any autonomous racing algorithm is desirable.  Because evaluating a neural network can be a computationally expensive task, our approach uses a multi-threading approach to limit the impact of the computation required to evaluate our model.  At runtime, a separate thread is spawned to maintain a circular buffer containing the $C$ images that make up the input to both the waypoint predictor and the B\`ezier Curve predictor. This thread listens for new images and adds them to the circular buffer, completely independent of the main control loop. The main control loop then grabs a snapshot of this buffer and runs the model to produce a control value.  This approach guarantees that the sequence of images in the circular buffer are \emph{always} contiguous in time, there is no risk of dropping an image because the main control loop is busy evaluating a neural network.  Empirical measurements of the runtimes of each of these loops are explored in subsection \ref{subsec:runtime}.

Additionally, to generate a lookahead point for the Pure Pursuit controller, the B\`ezier Curve predictor takes uniformly spaced samples from the predicted B\`ezier Curve and selects a lookahead point from these samples.  

To generate throttle commands, a bang-bang control approach is used.  If the car's current velocity is slower than the reference velocity of the predicted B\`ezier Curve at the lookahead point, the throttle is set to it's maximum value and brake is set to 0. Otherwise, the brake is set to it's maximum value and throttle is set to 0.  The Waypoint Predictor uses a similar approach, but fits a spline to it's predicted waypoints and evaluates that spline's derivative to generate a reference velocity.

These details are specific to our experiments, the models presented are scalable to other strategies for transforming a predicted trajectory into steering and throttle commands.

We implement these models in the PyTorch \citep{pytorch} framework.   Additionally, an NVIDIA GeForce GTX 1080Ti was used to test each model. Each model was trained on a distributed cluster containing many NVIDIA GPUs.

\section{Experimental Results}
\label{sec:experiments}
We now present several empirical results of this work. 
\begin{enumerate}
    \item An evaluation of how well our data-logging infrastructure is synchronized with the F1 game
    \item Open-loop results of several neural network architectures trained on data from the game
    \item A closed-loop driving evaluation of those same models
    \item An evaluation of the running time of our approach
\end{enumerate}

We show that our infrastructure infrastructure produces negligible error between logged values and the true internal state of the F1 game. Additionally we provide case studies of several deep learning models both in the open-loop (root-mean-square error) sense and in the closed-loop (true driving performance) sense.

\subsection{Data Logging Infrastructure}
\begin{figure}[!htb]
\centering
\includegraphics[width=0.65\textwidth]{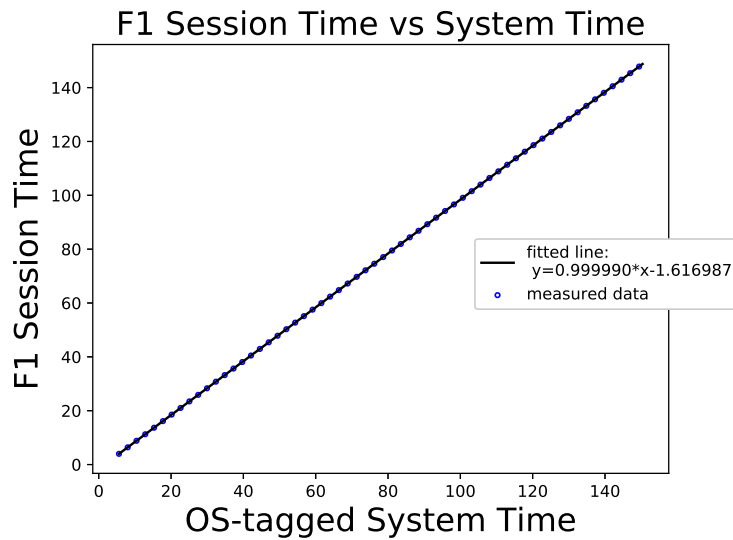}
\caption{A plot of F1 session times versus measured system times.  For readability, the data has been downsampled by a factor of 100. The slope of the regression line very close to $1.0$ implies that the timing between the F1 game clock and the measurable system clock is consistent.  The intercept of $-1.616876$ on the regression line indicates that the F1 session began broadcasting $1.616876$ seconds before the datalogger was started.}
\label{fig:datalogger_regression}
\end{figure}
We present an evaluation of our datalogging infrastructure by measuring the relative slope between the timescale on the measured OS clock and the received timestamps from the F1 game. In the ideal case, a slope of 1.0 would indicate that 1 second on our system clock corresponds to exactly 1 second on the games internal clock, indicating our data logging infrastructure is synchronized perfectly with the game. We evaluate our datalogging infrastructure by fitting a regression line to measurements of the system timestamp assigned to the UDP packets versus their stated F1 session time. Figure \ref{fig:datalogger_regression} shows a plot of such a regression line. The slope of this line is $0.99999$, indicating that each each second of system time adds $1$ microsecond of error in the corresponding session time.  To put this in perspective, a data logging session that lasts 4 hours would only introduce 14 milliseconds of error. NumPy reported an $r^2$ value of exactly $1.0000$, indicating that the error in this fit is so small, NumPy is numerically incapable of computing it. The intercept of $-1.616876$ on the fit line indicates that the measurements on the system clock began $1.616876$ seconds after the F1 session began.

\subsection{Neural Network Architectures}
Each neural network architecture was trained for 100 epochs under it's corresponding loss function with a mini-batch size of 128.  Stochastic Gradient Descent with a step size of ${10}^{-4}$ was used as the underlying optimization routine for network weight training.  For the B\'ezier Curve predictor, we use weighting factors of $w_{position}=1.0,w_{velocity}=0.1,w_{control\textunderscore point}=0.05$.  Each model was then tested on the Australia circuit by having each model control a standard F1 car on 5 test laps around the track, the car was reset to the same starting position on each lap.  To measure how effectively each model races a lap, we define a ``boundary failure" (BF) to be when an autonomous agent veers outside the bounds of the track.  We define a ``Boundary Failure Score" (BFS) to be the average distance outside the track the autonomous agent went during a boundary failure. A boundary failure with a lower BFS is considered a better (``less bad") failure.

\subsubsection{Open-Loop RMSE Results}

A traditional way to evaluate neural network models is with a classical ``training set/validation set" approach where models are trained on a specified dataset and then tested on an unseen validation set under some open-loop metric. 
We show that such a metric does not necessarily correlate with real-time driving performance.  
We performed such an evaluation of the purely end-to-end models. The Root-Mean-Square-Error (RMSE) between predicted steering and ground-truth steering is used as an evaluation metric.

\begin{table}[!htb]
\centering
\label{tab:my-table}
\begin{tabular}{|l|l|l|}
\hline
\textbf{Model Configurations} & \textbf{RMSE Steering} & \textbf{RMSE Throttle} \\ \hline
PilotNet & 0.179 & 0.317 \\ \hline
CNN-LSTM & 0.184 & 0.302 \\ \hline
\end{tabular}
\end{table}




Note that the CNN-LSTM setup actually performs \emph{worse} than PilotNet in the open-loop sense in predicted steering angles and slightly better than PilotNet in predicting throttle values.

\subsubsection{Closed Loop (Driving Performance) Results}
To more appropriately evaluate these neural network architectures, we use a fully closed-loop approach.  Rather than evaluating based only on an offline metric like Root-Mean-Square error. We utilize our DeepRacing framework's ability to close the loop and send steering, throttle, and brake commands back into the Codemaster's\textsuperscript{\texttrademark} game to evaluate each network on it's ability to actually race a simulated F1 car.  Specifically, we trained each model on {~}{$\sim$}25000 images of training data from the Australia F1 circuit.  For PilotNet and the CNN-LSTM architectures, each image is labeled with steering and acceleration.  For both the Waypoint predictor and the B\'ezier Curve predictor, each image is labeled with 60 future waypoints from an expert driver.

For these experiments, a context length of $C=5$ is used for both the waypoint predictor and the B\`ezier Curve predictor.  The waypoint predictor was configured to predict 20 timesteps into the future, corresponding to 1.4 seconds.  The same timescale was used for the B\`ezier Curve predictor. 

We ran each model 5 times and recorded the following metrics:

\begin{enumerate}
    \item Whether the model successfully completed a lap
    \item Mean Lap time (if a lap was successfully completed)
    \item Mean time between boundary failures (TBF)
    \item Mean distance along the track between boundary failures (DBF)
    \item Mean Number of Boundary Failures (NBF)
    \item Mean Boundary Failure Score (BFS)
\end{enumerate}

\begin{figure}[!htb]
    \centering
    \includegraphics{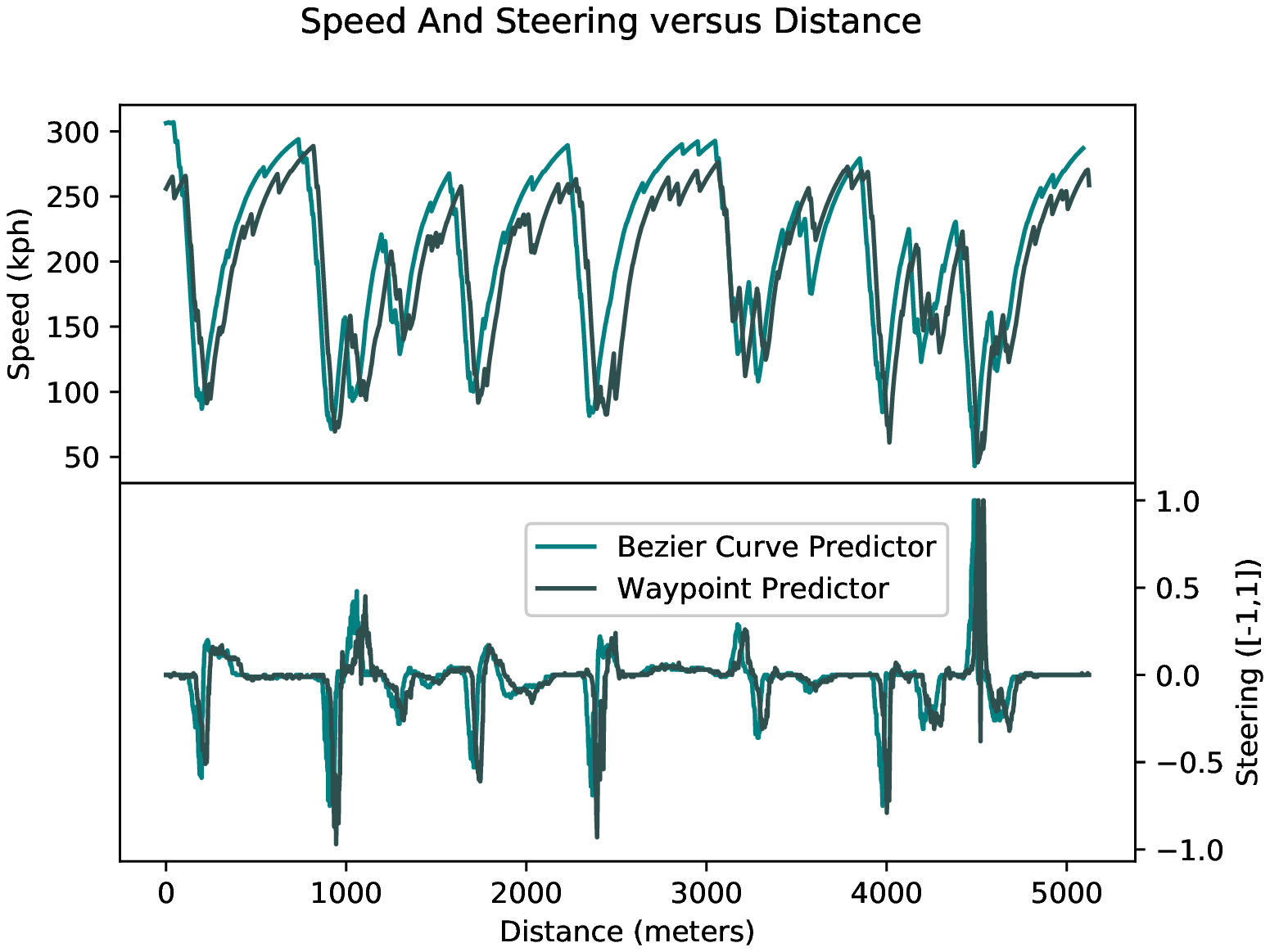}
    \caption{A Formula 1 style control plot for a test run. The B\'ezier Curve predictor produces much smoother velocity profiles}
\end{figure}
\begin{figure}[!htb]
    \centering
    \includegraphics[width=0.8\columnwidth]{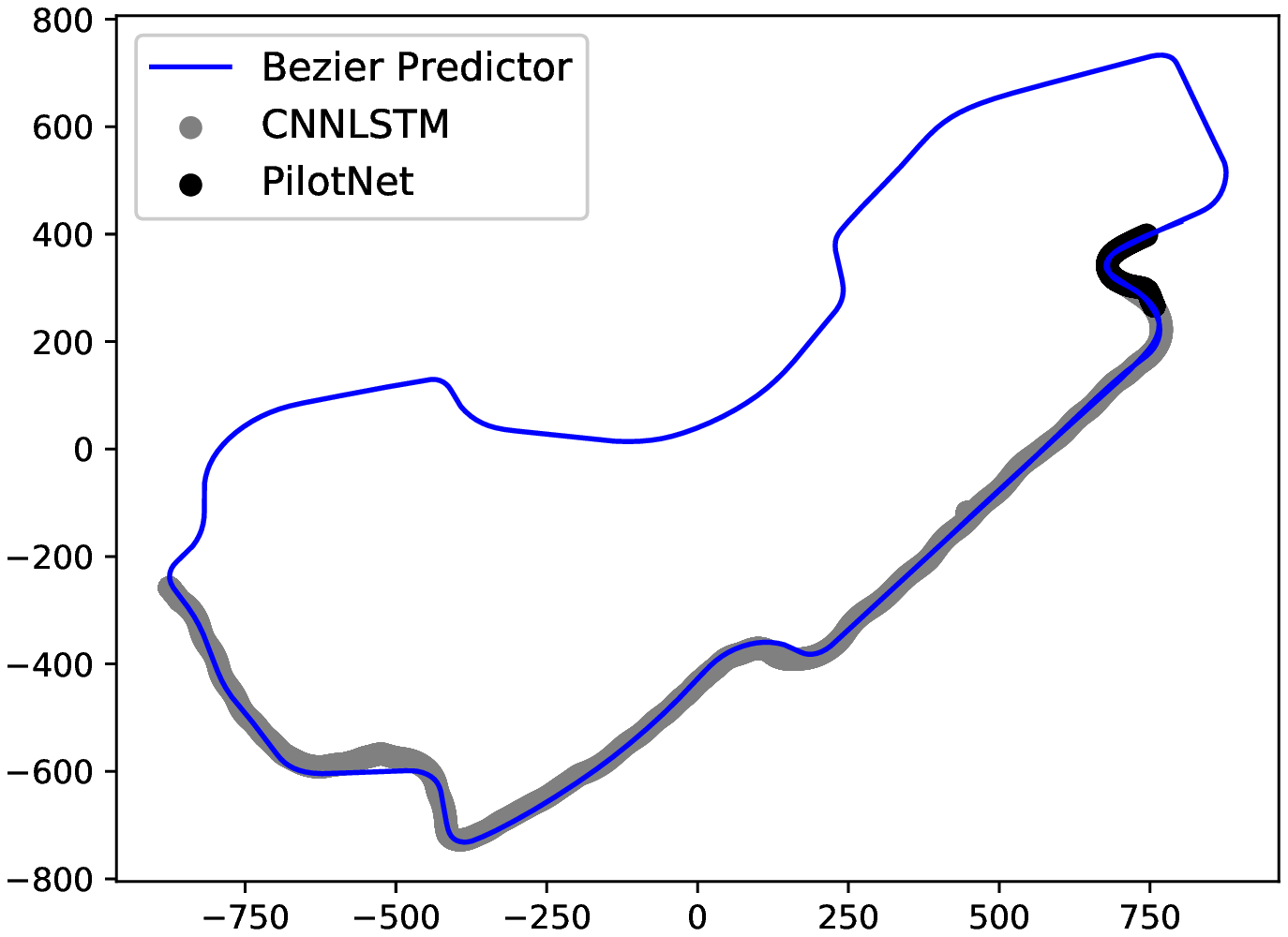}
    \caption{A plot of the path followed by our approach versus the two open-loop baselines. PilotNet fails almost immediately and the CNNLSTM only makes it about half-way around the track.}
    \label{fig:velocity_comparison}
\end{figure}
Additionally, we conduct an ablative analysis of the effect of using the 3D Convolutional layers to encode to compute the inputs for the final recurrent calls to the LSTM layer.  For comparison, we evaluate both the waypoint predictor and the B\'ezier Curve predictor with constant values as the input to the final LSTM calls.  These constant values are defined as model parameters and their values are learned during the training process. The results from these experiments are tabulated below.  All figures are arithmetic means across all 5 runs. DNF indicates ``did not finish" a successful lap.
\begin{table}[]
\centering
\resizebox{\textwidth}{!}{%
 \begin{tabular}{|c c c c c c c|} 
 \hline
 \makecell{Model\\Configuration} & \makecell{Lap Time\\(seconds)} & \makecell{TBF\\(seconds)} & \makecell{DBF\\(meters)} & \makecell{Number of \\Boundary Failures} & \makecell{BFS\\(meters)} & Successful Laps \\ 
 \hline
 PilotNet & DNF & 4.07 & 181.444 & 1.800 & 4.267 & 0 \\ 
 \hline
 CNN-LSTM & DNF & 6.367 & 304.490 & 3.600 &  2.539 & 0\\
 \hline
 \makecell{Waypoint Predictor\\ without 3D Convolution}& 113.38 & 11.92 & 626.23 & 6.6 & 7.02 & 1\\
 \hline
 Waypoint Predictor & 106.683 & 16.739 & 855.817 & 5.6 & 0.239 & 4\\
 \hline
 \makecell{B\'ezier Curve Predictor \\without 3D Convolution}& \makecell{$\mathbf{99.95}$} & 19.01 & 1008.46 & 7.4 & 2.89 & 5\\
 \hline
 B\'ezier Curve Predictor & 101.72 & \makecell{$\mathbf{33.62}$} & \makecell{$\mathbf{1786.36}$} & \makecell{$\mathbf{1.8}$} & \makecell{$\mathbf{0.041}$} & \makecell{$\mathbf{5}$}
 
\\ \hline
\end{tabular}
}
\caption{Results of our closed-loop testing. Note that the B\'ezier Curve Predictor outperforms all of the other models on (almost) all metrics. Also note that removing the 3d convolutional layers and replacing their outputs with learnable constants significantly degrades performance on both trajectory prediction models. ``DNF" indicates the model did not finish a lap.}
\label{table:closed_loop}
\end{table}

Note that the B\'ezier Curve predictor with the 3D convolution outperforms all 5 other models on all metrics, except for lap time.  However, the modest improvement in lap time by removing the 3D convolutional subnetwork comes at the expense of significantly more boundary failures with a significantly larger boundary failure score, indicating that this improvement in lap time is really just a result of illegally ``cutting corners".  PilotNet was unable to complete a successful lap, as it crashed into a wall almost immediately (within 5 seconds on all 5 runs). The CNNLSTM architecture was also unable to complete a successful lap, but managed to make it around the first turn and down the initial straightaway. The B\'ezier Curve predictor also results in a smoother velocity curve than direct waypoint regression (see Figure \ref{fig:velocity_comparison}).

Interestingly, the CNNLSTM architecture outperforms PilotNet in the closed loop sense (it was at least able to make the initial turn and most of the way down the initial straightaway) despite performing slightly worse than PilotNet under a purely open-loop evaluation.

Finally, we performed a longevity test of each of the trajectory-based methods. In this test, each model was deployed to a test run and allowed to continue until the vehicle either crashed or became inoperable.  The results of this longevity test are in table \ref{table:shake_to_break}.
\begin{table}[]
\centering
\begin{tabular}{|c|c|}
\hline
\begin{tabular}[c]{@{}c@{}}Model \\ Configuration\end{tabular}                       & \begin{tabular}[c]{@{}c@{}}Laps To\\ Failure\end{tabular} \\ \hline
Bézier Curve Predictor                                                               & 106                                                       \\ \hline
\begin{tabular}[c]{@{}c@{}}Bézier Curve Predictor\\ (No 3d Convolution)\end{tabular} & 86                                                        \\ \hline
Waypoint Predictor                                                                   & 11                                                        \\ \hline
\begin{tabular}[c]{@{}c@{}}Waypoint Predictor\\ (No 3d Convolution)\end{tabular}     & 1                                                         \\ \hline
\end{tabular}
\caption{The results of out longevity testing.  Both fully end-to-end models did not complete a single lap and are excluded from this table.}
\label{table:shake_to_break}
\end{table}

\newpage

\subsection{Computation Time Analysis}
\label{subsec:runtime}
The main control loop for each approach consists of three high level steps:
\begin{enumerate}
    \item Get a snapshot of the circular buffer containing the $C$ most recently measured images
    \item Evaluate the model on this snapshot
    \item Generate steering/throttle commands from the model's output
\end{enumerate}

We measured the running time of this control loop for each approach.  Additionally, we measure the running time required to save a newly received image to the circular buffer.  The results are tabulated in table \ref{table:runtime_analysis}.
\begin{table}[]
\centering
\begin{tabular}{ccc}
\hline
\multicolumn{1}{|c|}{\begin{tabular}[c]{@{}c@{}}Model \\ Configuration\end{tabular}}                       & \multicolumn{1}{c|}{\begin{tabular}[c]{@{}c@{}}Mean\\ Runtime\\ (seconds)\end{tabular}} & \multicolumn{1}{c|}{\begin{tabular}[c]{@{}c@{}}Mean\\ Frequency\\ (Hz)\end{tabular}} \\ \hline
\multicolumn{1}{|c|}{Image Buffer}                                                                         & \multicolumn{1}{c|}{0.000707}                                                           & \multicolumn{1}{l|}{1,414.42}                                                        \\ \hline
\multicolumn{1}{|c|}{PilotNet}                                                                             & \multicolumn{1}{c|}{0.008}                                                              & \multicolumn{1}{c|}{125}                                                             \\ \hline
\multicolumn{1}{|c|}{CNNLSTM}                                                                              & \multicolumn{1}{c|}{0.011}                                                              & \multicolumn{1}{c|}{90.90}                                                           \\ \hline
\multicolumn{1}{|c|}{Waypoint Predictor}                                                                   & \multicolumn{1}{c|}{0.040}                                                              & \multicolumn{1}{c|}{25}                                                              \\ \hline
\multicolumn{1}{|c|}{\begin{tabular}[c]{@{}c@{}}Waypoint Predictor\\ (No 3d Convolution)\end{tabular}}     & \multicolumn{1}{c|}{0.0370}                                                             & \multicolumn{1}{c|}{27.034}                                                          \\ \hline
\multicolumn{1}{|c|}{Bézier Curve Predictor}                                                               & \multicolumn{1}{c|}{0.061}                                                              & \multicolumn{1}{c|}{16.396}                                                          \\ \hline
\multicolumn{1}{|c|}{\begin{tabular}[c]{@{}c@{}}Bézier Curve Predictor\\ (No 3d Convolution)\end{tabular}} & \multicolumn{1}{c|}{0.056}                                                              & \multicolumn{1}{c|}{17.816}                                                          \\ \hline
                                                                                                           &                                                                                         &                                                                             
\end{tabular}      
\caption{The runtime of each model as well as the thread for maintaining the image buffer.  The Bezier Curve Predictor is the slowest model, but is still sufficiently fast for an autonomous driving task.  Also note that the image buffer thread operates an order of magnitude faster than any of the tested models, indicating the risk of dropping an image due to computational overhead is very low.}
\label{table:runtime_analysis}  
\end{table}

The B\`ezier Curve predictor is the slowest model, this is not surprising as it involves the most neural network layers. However, at $16.393 Hz$, it is still sufficiently fast for autonomous racing.  Additionally, note that the separate thread maintaining the image buffer runs at over $1414Hz$, several orders of magnitude faster than the image capture rate of our infrastructure, indicating the risk that this approach will drop images and produce discontinuous image sequences is negligible.


\section{Conclusion and Future Work}
\label{sec:conclusion}

\subsection{Discussion}
In this paper, we address the problem of learning expert agile driving behavior from demonstration. We consider motor-sport racing as the proxy for agile driving since it involved high speed driving at the limits of control, and often in close proximity of other vehicles. This hypothesis leads us to the problem of learning how to autonomous race in a realistic motor-sport racing environment. 
One of our primary contribution is DeepRacing - a novel end-to-end framework, and a virtual testbed for training and evaluating algorithms for the hard challenge of autonomous racing.  
The virtual testbed is implemented using a highly realistic and professional Formula One gaming environment.  
Our open-source DeepRacing framework, which is integrated with ROS2, will enable researchers to explore the limits of vision based end-to-end autonomous racing. 

In addition, we also develop and present a new parametrized trajectory-based end-to-end trainable network for autonomous racing. Our novel AdmiralNet architecture, which uses B\'ezier Curves to parameterize a lower-dimensional manifold within the task space of the ego vehicle, outperforms traditional end-to-end networks by a very large margin and outperforms waypoint-based planning by $\sim5$ seconds in terms of lap time and by over 100\% in terms of time between failures, all while being robust and computationally tractable. 

\subsection{Future Work}
The results and the work so far has focused on autonomous racing in a time-trial manner, i.e. only one vehicle on the track at a time. Our ongoing and future work include a more focused study of head-to-head racing (multi-agent setting) with a stronger focus on real-world racing metrics like lap time and overall race position. We also intend to explore and compare our method with reinforcement learning based approaches for this problem.
Additionally, this work does not address how the agile behavior learned in a racing environment can make day-to-day driving safer. This challenging question of transferring agile behavior between domains is the subject of our future research in this area.  



\end{document}